\documentclass[runningheads]{llncs}

\usepackage[mobile]{eccv}

\usepackage{eccvabbrv}

\usepackage{graphicx}
\usepackage{booktabs}

\usepackage[accsupp]{axessibility}  

\usepackage{makecell}
\usepackage{multirow}
\usepackage[table]{xcolor}
\usepackage{pifont}
\usepackage{arydshln}

\newcommand{\cmark}{\ding{51}}
\newcommand{\xmark}{\ding{55}}
\definecolor{firstcolor}{HTML}{E2F0D9}
\definecolor{secondcolor}{HTML}{FFF2CC}
\definecolor{thirdcolor}{HTML}{DDEBF7}

\usepackage{hyperref}

\usepackage{orcidlink}

\makeatletter
\def\@fnsymbol#1{\ensuremath{%
    \ifcase#1
    \or \dagger
    \or \ddagger
    \or \mathsection
    \or \mathparagraph
    \else \@ctrerr
    \fi}}
\makeatother

\begin{document}

\title{SAF3R: Dynamic Sparse Attention for Feed-Forward 3D Reconstruction Transformers} 

\titlerunning{SAF3R}

\author{
Jianing Deng\inst{1}\orcidlink{0000-0001-7307-6629} \and
Yuanzhe Li\inst{2}\orcidlink{0009-0001-4633-0641} \and
Jialu Wang\inst{3}\orcidlink{0009-0009-4671-7430} \and
Song Wang\inst{4}\orcidlink{0000-0003-1273-7694} \and \\ 
Tianlong Chen\inst{5}\orcidlink{0000-0001-7774-8197} \and
Huanrui Yang\inst{2}\orcidlink{0000-0002-3384-4512} \and
Jingtong Hu\inst{1}\thanks{Corresponding author.}\orcidlink{0000-0003-4029-4034}
}

\authorrunning{J.~Deng et al.}

\institute{
University of Pittsburgh, Pittsburgh, PA, USA \and
University of Arizona, Tucson, AZ, USA \and
Tongji University, Shanghai, China \and
University of Central Florida, Orlando, FL, USA \and
University of North Carolina at Chapel Hill, Chapel Hill, NC, USA \\
\email{\{jid70,jthu\}@pitt.edu}
}

\maketitle

\begin{abstract}
  Feed-forward 3D reconstruction (F3R) transformers have recently achieved remarkable success. However, scaling them to long image sequences remains challenging, as the quadratic complexity of cross-view global attention quickly becomes the dominant computational bottleneck. While recent efforts attempt to improve efficiency through compressed or sparse attention, they fail to fully exploit the inherent sparsity and dynamic behavior of global attention. In this work, we present a comprehensive analysis of global attention across multiple F3R transformers and reveal that attention patterns are highly heterogeneous, dynamic, and extremely sparse across layers and attention heads. Motivated by these findings, we propose SAF3R, a training-free dynamic sparse attention framework tailored to F3R transformers. SAF3R integrates tailored sparse attention mechanisms with offline head profiling and an efficient online adaptation strategy to match input-dependent attention behaviors. Extensive experiments demonstrate that SAF3R achieves high sparsity ratios while preserving camera pose estimation and 3D reconstruction quality, translating into substantial end-to-end speedup on F3R transformers compared to existing methods. Code is available at \url{https://github.com/jndeng/SAF3R}.
    
  \keywords{3D Reconstruction \and Transformers \and Sparse Attention}
\end{abstract}

\section{Introduction}
\label{sec:intro}

\begin{figure}[tb]
  \centering
  \includegraphics[width=\textwidth]{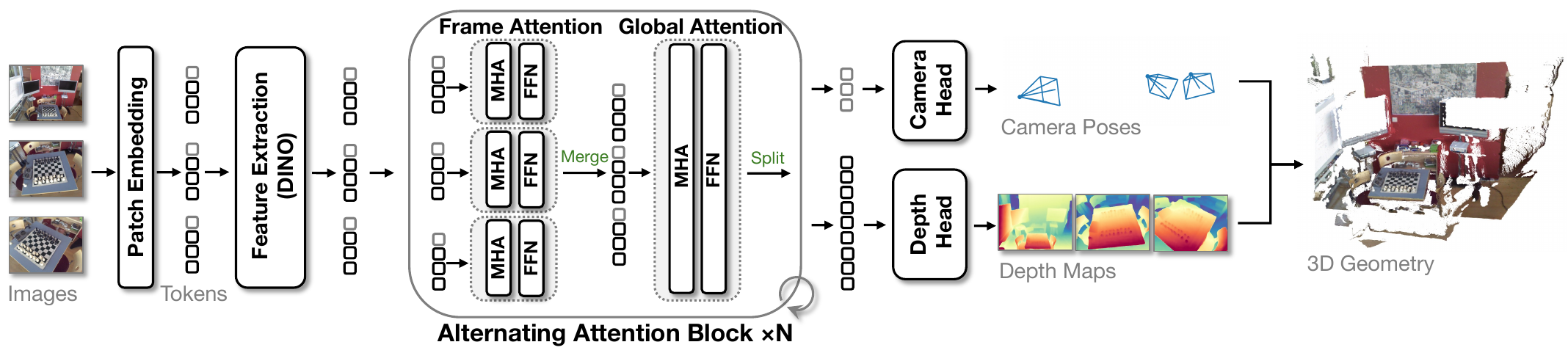}
  \caption{Overview of the architecture shared by existing F3R transformers~\cite{wang2025vggt,keetha2025mapanything,wang2025pi,lin2025depth}. Image patches are first encoded by a DINO~\cite{caron2021emerging,oquab2023dinov2} encoder and then processed by $N$ alternating attention blocks of local and global attention, each consisting of multi-head self-attention (MHA) and feed-forward networks (FFN).
  }
  \label{fig:f3r arch}
\end{figure}

Recent advances in deep learning have catalyzed a paradigm shift in 3D geometric estimation, transitioning from iterative optimization-based pipelines~\cite{shotton2013scene,schonberger2016structure,furukawa2015multi,schonberger2016pixelwise} to end-to-end neural architectures that directly infer geometry from raw images~\cite{wang2024dust3r,leroy2024grounding,yang2025fast3r,wang2024vggsfm,yang2024depth}. In particular, large-scale feed-forward 3D reconstruction (F3R) models~\cite{wang2025vggt,wang2025pi,keetha2025mapanything,lin2025depth,wang2026vggto} have achieved remarkable progress in geometric reconstruction and multi-view scene understanding. 
These models adopt a unified transformer architecture~\cite{vaswani2017attention} that jointly performs multiple geometric tasks, including camera pose estimation and dense point map prediction. A key characteristic of F3R transformers is the use of cross-view multi-head self-attention~\cite{vaswani2017attention,dosovitskiy2020image} to model long-range geometric relationships across frames. By allowing tokens from all views to interact globally, they directly infer consistent scene geometry and camera poses in an end-to-end feed-forward manner, eliminating the need for explicit feature matching or global optimization. \cref{fig:f3r arch} illustrates the general architecture shared by existing F3R transformers~\cite{wang2025vggt,wang2025pi,keetha2025mapanything,lin2025depth}.

\begin{figure}[tb]
  \centering
  \includegraphics[width=\textwidth]{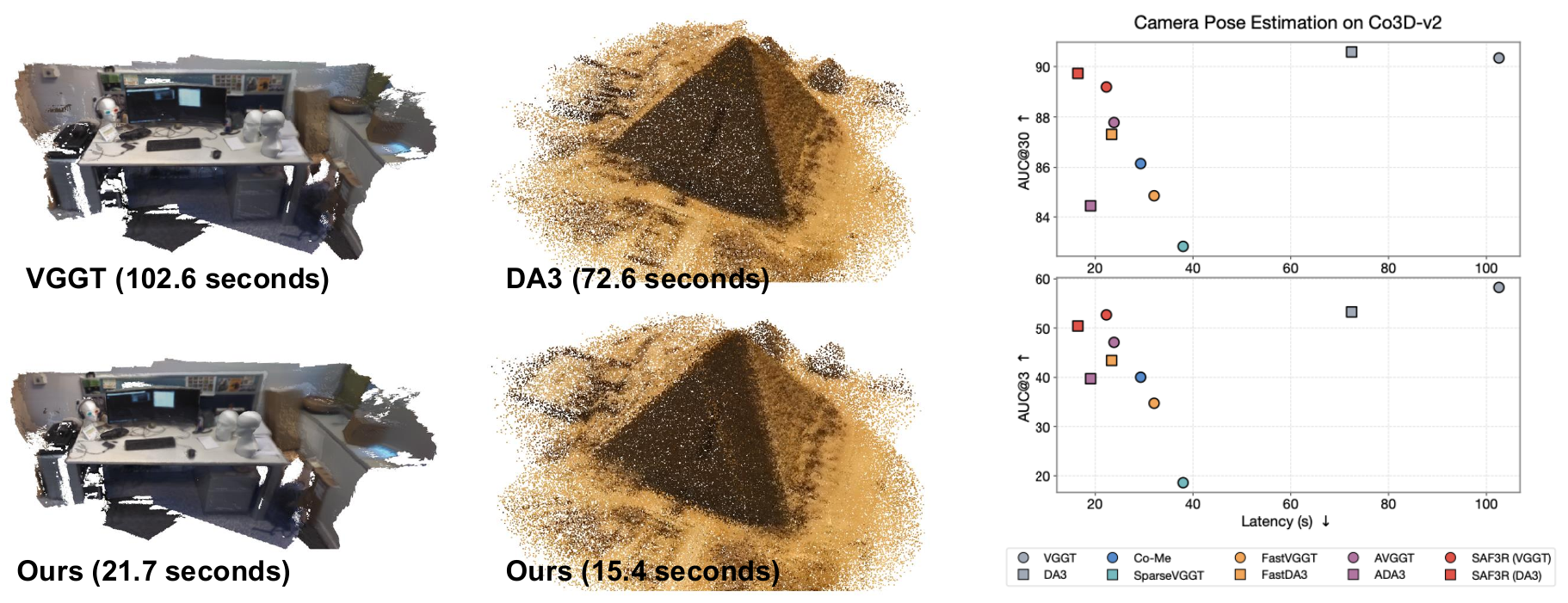}
  \caption{\textbf{Left:} SAF3R accelerates feed-forward 3D reconstruction (F3R) models while preserving reconstruction quality. \textbf{Right:} SAF3R outperforms prior efficient F3R methods on camera pose estimation and achieves performance close to the original model.
  }
  \label{fig:teaser}
\end{figure}

However, scaling F3R transformers to long image sequences remains challenging, as the quadratic complexity of cross-view global attention quickly dominates memory consumption and latency, severely limiting scalability and real-time deployment~\cite{shen2025fastvggt,wang2025faster}.
To mitigate this issue, recent works explore efficient approximations to full global attention, including token compression~\cite{shen2025fastvggt,wang2025flashvggt,chen2025co,shu2025litevggt,wang2025httm} and sparse attention mechanisms~\cite{wang2025faster,ren2026speed3r,sun2025avggt}. A summary of these methods is provided in~\cref{tab:method comparison}. While these approaches reduce computational cost, they typically adopt static or homogeneous sparsity configurations shared across layers and heads. Such designs overlook a critical property of F3R transformers: their unified multi-task architecture, which jointly models sparse camera pose estimation and dense point cloud reconstruction, induces inherently non-uniform and dynamic global attention patterns. Instead, our analysis reveals that global attention behavior varies significantly across layers, heads, and input scenes, suggesting that naïve sparsification may discard critical cross-view geometric information or waste computation on uninformative attention links.

Based on these insights, we introduce SAF3R, a training-free dynamic sparse attention inference framework tailored to F3R transformers.
The SAF3R framework determines and applies the most appropriate sparse attention pattern for \emph{each individual head} across all global attention layers, fully leveraging the heterogeneous and input-dependent nature of global attention.
Through comprehensive analysis, we first categorize global attention heads into four types and design tailored sparse attention kernels for each. Building on this design, SAF3R adopts a two-stage pipeline consisting of offline head profiling and online pattern adaptation. In the offline stage, we search for head-specific sparse configurations starting from a uniform-stride sparse attention baseline~\cite{shen2025fastvggt,sun2025avggt}, progressively refining it while ensuring no increase in approximation error on a calibration dataset. During inference, SAF3R executes these optimized configurations and dynamically adjusts input-dependent sparse patterns (\eg, Top-$K$ indices) through lightweight approximation and index caching. We evaluate SAF3R on multiple F3R models and benchmarks.
As shown in~\cref{fig:teaser}, our method maintains both camera pose estimation and point cloud reconstruction quality while providing up to 7$\times$ end-to-end speedup on long input sequences.

In summary, our contributions are three-fold:
(1) We conduct a comprehensive head-wise analysis of global attention in multiple F3R models, revealing heterogeneous and dynamic sparsity patterns across layers and heads.
(2) We propose SAF3R, a training-free dynamic sparse attention framework that exploits head-specific sparsity patterns to accelerate model inference.
(3) We demonstrate significant computational savings and substantial inference speedup on multiple F3R models while maintaining camera pose and 3D reconstruction accuracy.

\section{Related Work}
\label{sec:related work}

\subsection{Feed-forward 3D Reconstruction Transformers}

Early 3D reconstruction pipelines rely on multi-stage optimization methods such as Structure-from-Motion (SfM)~\cite{schonberger2016structure} and Multi-View Stereo (MVS)~\cite{furukawa2015multi}, which have been widely adopted in practice~\cite{schonberger2016structure,schonberger2016pixelwise,mur2017orb} but are computationally expensive and prone to error accumulation. Recent advances in deep learning have enabled feed-forward 3D reconstruction (F3R) models that directly infer scene geometry from unseen images. Early approaches~\cite{wang2024dust3r,leroy2024grounding,wang2024vggsfm,yang2025fast3r} established the paradigm of generalizable 3D reconstruction from uncalibrated images. More recent models~\cite{wang2025vggt,wang2025pi,keetha2025mapanything,lin2025depth,fang2026dens3r,gao2026more,wang2026vggto}, including VGGT~\cite{wang2025vggt}, $\pi^3$~\cite{wang2025pi}, and Depth Anything 3~\cite{lin2025depth},  further improve unified multi-view geometric reasoning, permutation invariance, metric scale recovery, and reconstruction quality. A common design among these offline F3R models is to jointly process the available input views using global self-attention to capture long-range cross-view geometric relationships.
In parallel, recent works have explored streaming reconstruction, where observations are processed incrementally as they arrive using recurrent states, causal attention, or explicit geometric memories~\cite{wang2024cut3r,chen2026ttt3r,zhuo2026streaming,lan2026stream3r,wu2026point3r,wang2026stac,zhang2026loger,yu2026towards,chen2026geometric,cheng2026horizonstream}.
In this paper, we focus on the \emph{offline} F3R setting, where all input views are available during inference and can be jointly reasoned over. While global attention enables effective cross-view geometric reasoning, its quadratic computational complexity and growing memory footprint become major bottlenecks as the number of input views increases.

\subsection{Leveraging Sparsity in Attention}
Sparse attention has emerged as an effective strategy to alleviate the quadratic computational cost of attention-based architectures across various domains~\cite{zhang2025spargeattn,zhang2025fast,xiao2023efficient,xiao2024duoattention,xu2025xattention,xi2025sparse,yang2025sparse,yuan2025native,jiang2024minference,lai2025flexprefill}.
SpargeAttn~\cite{zhang2025spargeattn} uses a low-resolution attention approximation to identify important blocks and computes attention only on them. DuoAttention~\cite{xiao2024duoattention} identifies two general sparse attention patterns in large language models and learns to assign each head to one of them using synthetic calibration data.
NSA~\cite{yuan2025native} trains the model to combine several predefined sparse attention patterns. MInference~\cite{jiang2024minference} uses offline profiling to assign head-specific sparse attention patterns for efficient inference. FlexPrefill~\cite{lai2025flexprefill} and SVG~\cite{xi2025sparse} use online profiling to determine input- or head-specific sparse attention patterns.
However, unlike the relatively stable attention patterns in language models or the spatio-temporal locality in vision transformers, F3R models exhibit highly heterogeneous and task-dependent attention distributions across layers and heads due to their multi-task nature. This motivates systematic head-level analysis and tailored sparse kernel designs for F3R transformers.

\begin{table}[t]
  \caption{
  Comparison of efficient global attention mechanisms for F3R transformers.
  }
  \label{tab:method comparison}
  \centering
  \scriptsize
  \setlength{\tabcolsep}{7pt}
  \begin{tabular}{@{}lccccc@{}} 
    \toprule
    \textbf{Method} 
    & \makecell{\textbf{Compression}\\\textbf{Strategy}} 
    & \makecell{\textbf{Policy}\\\textbf{Scope}} 
    & \makecell{\textbf{Input-}\\\textbf{adaptive}} 
    & \textbf{Head-wise} 
    & \makecell{\textbf{Dynamic}\\\textbf{Sparsity}} \\ 
    \midrule
    FastVGGT~\cite{shen2025fastvggt}   
        & Token Merging   & Model-level & \xmark & \xmark & \xmark \\
    FlashVGGT~\cite{wang2025flashvggt} 
        & Token Merging   & Model-level & \cmark & \xmark & \xmark \\
    Co-Me~\cite{chen2025co}      
        & Token Merging   & Model-level & \cmark & \xmark & \xmark \\
    LiteVGGT~\cite{shu2025litevggt}   
        & Token Merging   & Model-level & \cmark & \xmark & \xmark \\
    HTTM~\cite{wang2025httm}       
        & Token Merging   & Model-level  & \xmark & \cmark & \xmark \\
    SparseVGGT~\cite{wang2025faster} 
        & Sparse Attention & Model-level & \cmark & \xmark & \xmark \\
    Speed3R~\cite{ren2026speed3r}
        & Sparse Attention & Model-level & \cmark & \xmark & \xmark \\
    AVGGT~\cite{sun2025avggt}      
        & Sparse Attention & Stage-level & \xmark & \xmark & \xmark \\
    \midrule
    SAF3R (Ours)      
        & Sparse Attention & Head-level & \cmark & \cmark & \cmark \\
    \bottomrule
  \end{tabular}
\end{table}

\subsection{Efficient Global Attention for F3R Transformers}

Recent works explore token and context compression to alleviate the computational bottleneck of global attention. FastVGGT~\cite{shen2025fastvggt} adopts token merging~\cite{bolya2022token,bolya2023token} to reduce the number of tokens before attention based on patch-level similarity. Subsequent works improve token merging with better anchor token selection. FlashVGGT~\cite{wang2025flashvggt} computes global descriptors and performs token merging via interpolation rather than simple averaging. Co-Me~\cite{chen2025co} uses confidence maps from F3R models to guide merging, while LiteVGGT~\cite{shu2025litevggt} further exploits geometric cues, such as edges, to select important anchor tokens. HTTM~\cite{wang2025httm} introduces head-wise token merging and spatio-temporal block grouping to reduce similarity-computation overhead. Beyond explicit token merging, recent works~\cite{jin2026zipmap,elflein2026vgg,xie2026scal3r} compress long-range scene context into compact neural representations through test-time adaptation, enabling scalable reconstruction over large image collections. These compression-based approaches reduce the effective context size, but either depend on token/anchor selection or require additional test-time adaptation.
On the other hand, SparseVGGT~\cite{wang2025faster} and AVGGT~\cite{sun2025avggt} approximate global attention with structured sparsity, yet their use of simple predefined sparsity patterns fails to model heterogeneous attention distributions. Speed3R~\cite{ren2026speed3r} introduces trainable sparse attention to approximate global attention, but requires costly model retraining.
As shown in~\cref{tab:method comparison}, existing token merging and sparse attention approaches do not jointly capture the dynamic and head-wise heterogeneous behavior of global attention, limiting their ability to fully exploit attention sparsity.

\section{Analysis of Global Attention in F3R Transformers}
\label{sec:analysis}

In this section, we conduct a comprehensive study of cross-view global attention in four representative F3R models: VGGT~\cite{wang2025vggt}, $\pi^3$~\cite{wang2025pi}, MapAnything~\cite{keetha2025mapanything}, and Depth Anything 3 (DA3)~\cite{lin2025depth}. Since attention is computed in a multi-head manner, we perform head-level analysis to capture heterogeneous behaviors across heads.
We use randomly sampled image sequences from the 7Scenes dataset~\cite{shotton2013scene}.
Below, we summarize our observations, which provide key insights into the underlying mechanisms, sparsity characteristics, and opportunities for acceleration.

\begin{figure}[t]
  \centering
  \includegraphics[width=\textwidth]{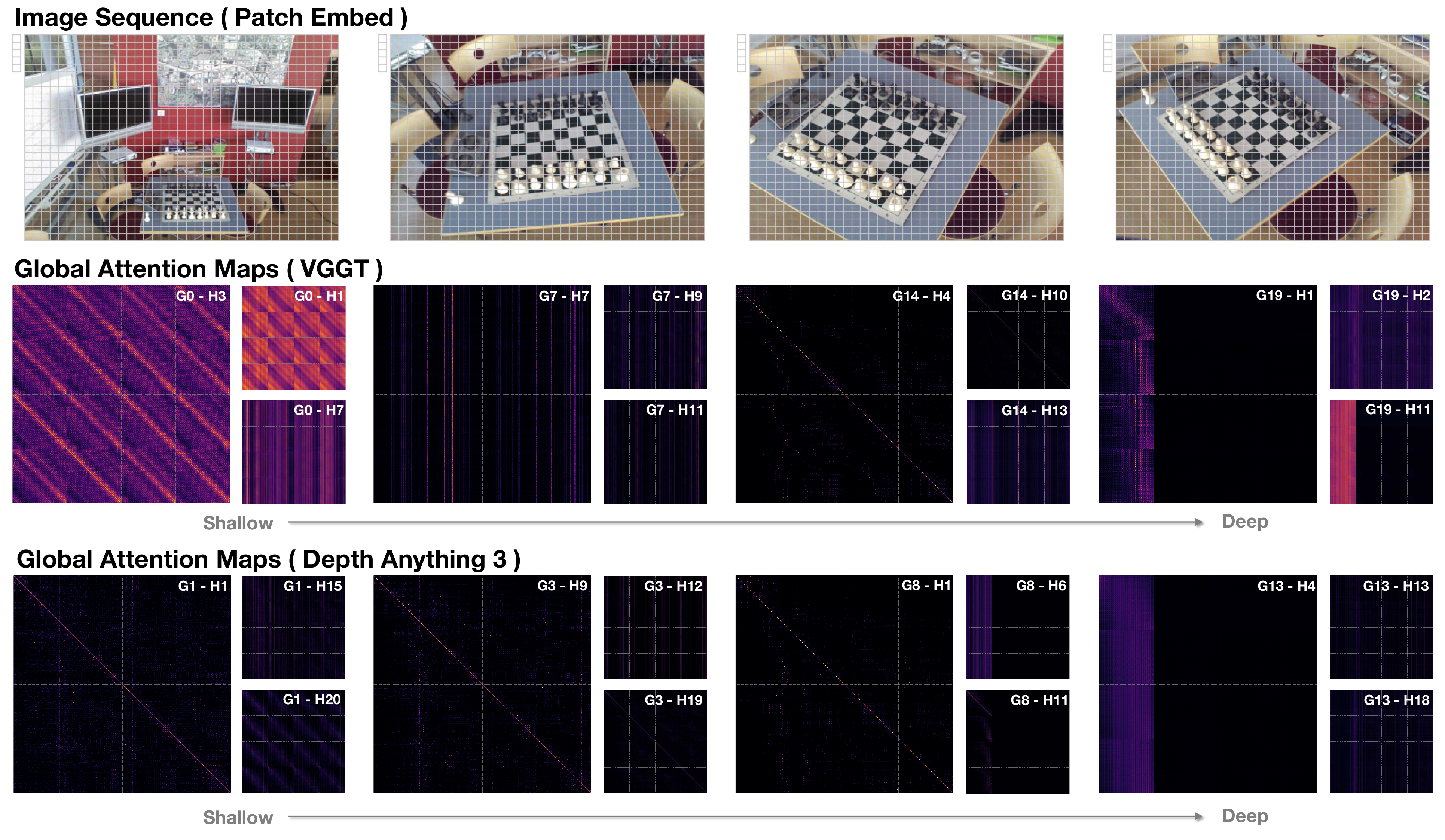}
  \caption{Representative global attention patterns across layers (G) and heads (H). Gray dashed lines in the attention maps separate tokens from different images. Similar observations apply to MapAnything and $\pi^3$ (see the appendix). Zoom in for best view.
  }
  \label{fig:attn pat vis}
\end{figure}

\textbf{Observation\#1: Global attention patterns are heterogeneous across models, layers, and heads.}
\cref{fig:attn pat vis} visualizes global attention maps across layers and heads in different F3R models. We observe a clear three-stage progression with distinct functional roles and sparsity patterns. In shallow layers, attention mainly performs local aggregation or focuses on a few salient patches, showing concentrated and often static sparsity with limited cross-view interaction. In middle layers, attention establishes point-to-point correspondences across views, where sparsity remains high but becomes semantically structured and content-dependent. In later layers, attention becomes more diffuse and mainly refines representations and camera poses relative to the anchor frame. This stage-wise behavior is consistent with prior findings~\cite{shen2025fastvggt,wang2025faster,sun2025avggt}.

\begin{figure}[t]
  \centering
  \includegraphics[width=\textwidth]{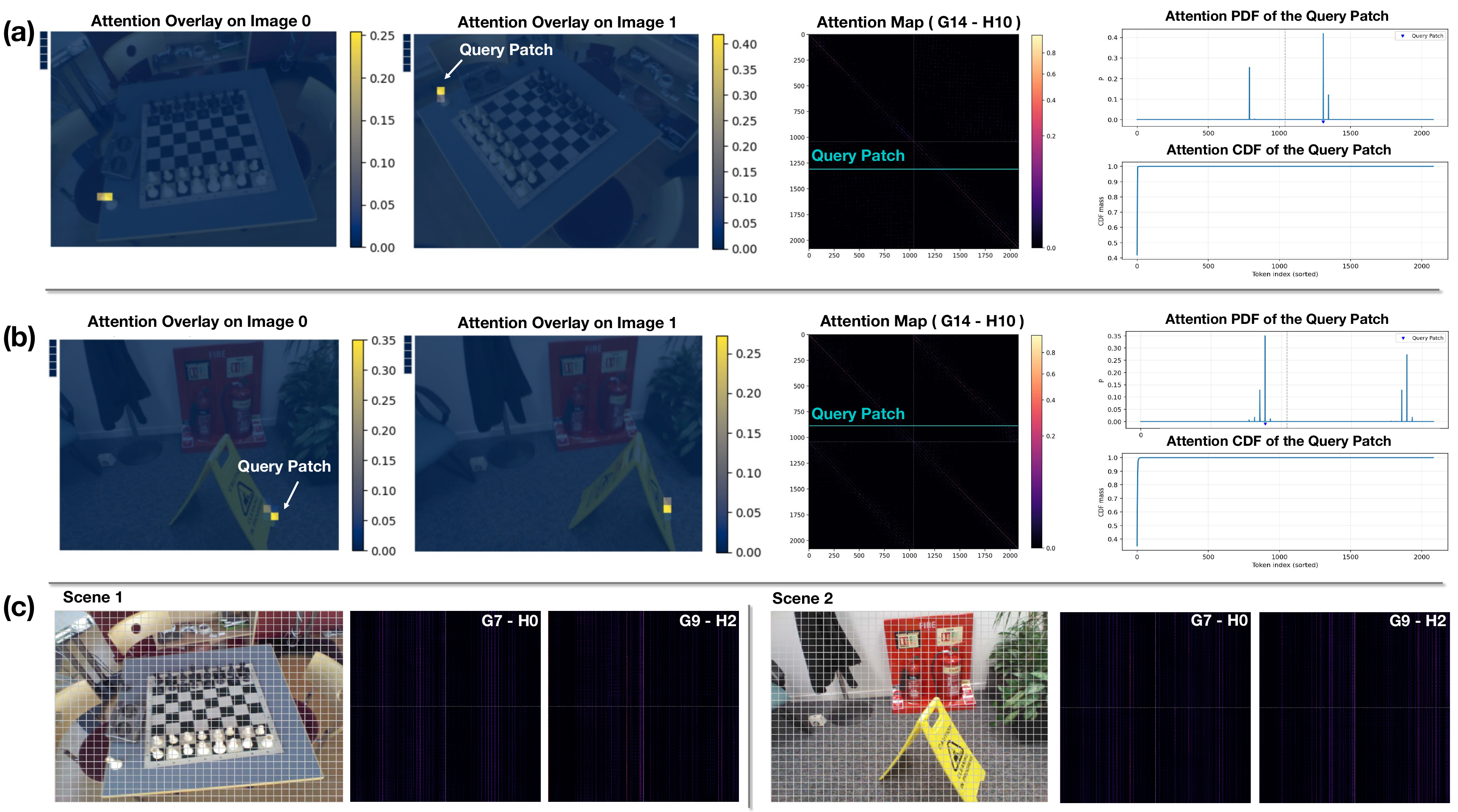}
  \caption{(a) and (b) illustrate the high sparsity of correspondence heads, while (c) shows the content-dependent dynamic nature of vertical-line heads.
  }
  \label{fig:dynamic heads}
\end{figure}

Beyond this stage-wise trend, global attention exhibits more nuanced heterogeneity. Attention patterns vary substantially \emph{across heads}, even within the same layer, suggesting that functional roles cannot be fully characterized at the coarse layer or stage level, as commonly assumed in prior work~\cite{shen2025fastvggt,wang2025faster,shu2025litevggt,sun2025avggt}. Moreover, attention behaviors differ across F3R models. For example, DA3 does not show the positional-encoding-induced local attention patterns observed in the early and late layers of VGGT. These findings highlight the need to model heterogeneity across both layers and heads, motivating a head-wise heterogeneous compression strategy for F3R transformers.

\textbf{Observation\#2: Attention heads can be categorized into four types based on their behaviors.} 
(1) \textbf{Position heads}, whose attention is dominated by positional encoding rather than semantic features, exhibiting patterns that are either fixed across all image blocks (\eg, G0-H3 and G0-H1) or predominantly attend to the anchor image which defines the camera coordinate frame (\eg, G19-H1 and G19-H11 in VGGT); 
(2) \textbf{Vertical-line heads}, where most queries consistently attend to a few critical keys (salient image patch tokens or sink tokens~\cite{darcet2023vision,xiao2023efficient}), forming prominent ``lines'' in the attention maps (\eg, G7-H7 and G14-H4); (3) \textbf{Correspondence heads}, where attention is primarily driven by semantic features, and each query selectively attends to the key corresponding to the same 3D location as shown in~\cref{fig:dynamic heads} (\eg, G14-H4 and G14-H10); (4) \textbf{Scanning heads}, which constitute the majority of attention heads, exhibit diverse attention patterns distributed across positions in all images.
The head-level diversity also induces distinct sparsity behaviors across head types and layers. Position and scanning heads are typically less sparse, while correspondence heads are highly sparse, attending to only a few key tokens. Vertical-line heads exhibit layer-dependent sparsity, with early-layer heads being sparser than middle ones.

\textbf{Observation\#3: Some global attention heads are highly dynamic (\ie, input-dependent).}
We observe that global attention heads exhibit both static and dynamic behaviors. 
While some heads produce attention patterns that remain largely consistent across inputs, others vary depending on the visual content. 
\cref{fig:dynamic heads} visualizes attention maps of two types of dynamic heads. 
As shown, these heads follow a fixed attention \emph{type} (\eg, capturing correspondence relationships or attending to a few specific keys), while the exact attention patterns remain input-dependent. This suggests that sparse attention patterns should not be fixed, but should instead adapt dynamically to the input.

\section{Proposed Dynamic Sparse Attention Framework}
\label{sec:method}

\begin{figure}[t]
  \centering
  \includegraphics[width=\textwidth]{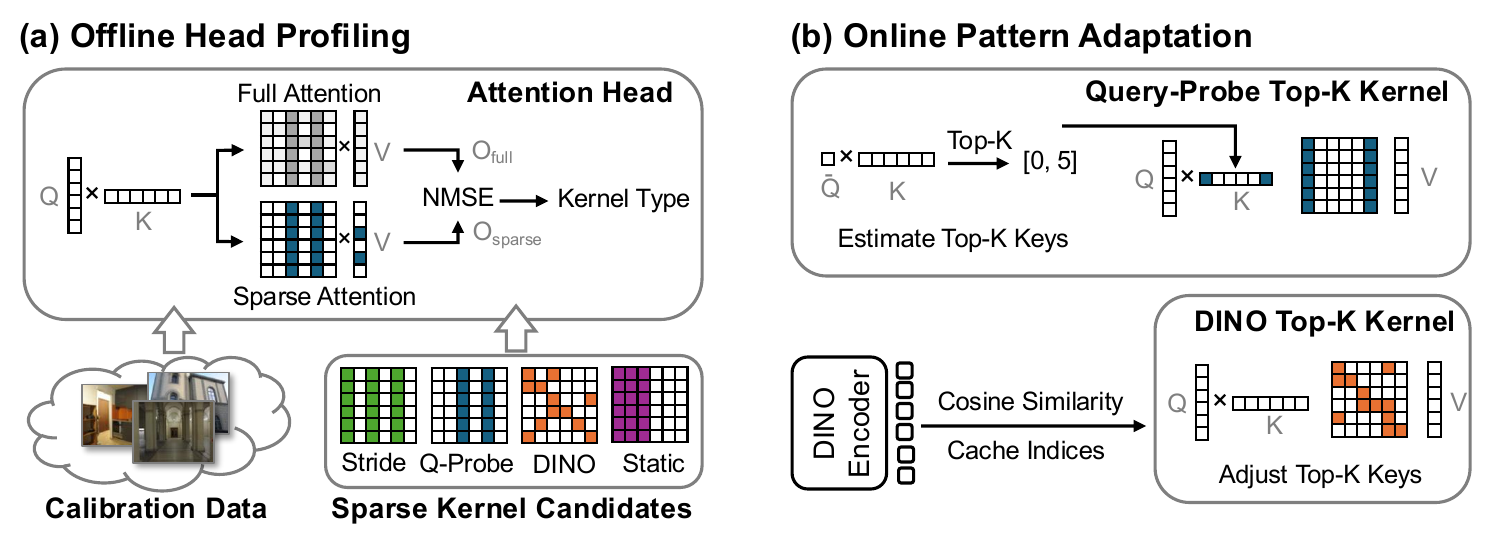}
  \caption{Overview of the proposed SAF3R framework.
  }
  \label{fig:saf3r overview}
\end{figure}

We present SAF3R, a two-stage dynamic \underline{S}parse \underline{A}ttention framework that exploits heterogeneous sparsity in \underline{F3R} transformers for efficient inference.
\cref{fig:saf3r overview} illustrates the overall framework.
We first introduce sparse attention mechanisms tailored to head-wise attention patterns (\cref{sec:custom sparse attn}). We then present an offline profiling strategy to identify the appropriate sparse type for each head (\cref{sec:offline profile}), followed by an online adaptation mechanism that dynamically adjusts attention patterns based on head types and input content during inference (\cref{sec:online adjustment}).

\subsection{Tailored Sparse Attention Kernels}
\label{sec:custom sparse attn}

Based on the head categorization in~\cref{sec:analysis}, we develop tailored sparse attention kernels for each attention type to exploit their distinct sparsity patterns. Each kernel reduces the computational complexity of attention from $\mathcal{O}(N^2)$ to $\mathcal{O}(N)$.

\textbf{Static Kernel.}
Since attention in positional heads is content-independent, we design a static kernel that computes fixed sparse attention patterns, \eg, allowing all queries to attend to a designated anchor frame or broadcasting the first frame's local attention map to other frames for cross-view interaction.

\textbf{Query-Probe Top-$K$ Kernel.}
For vertical-line heads, where attention mass is concentrated on a few critical key tokens (forming several prominent lines in the attention map), we restrict computation to attention scores between all queries and these key tokens, omitting the remaining interactions. As analyzed in~\cref{sec:analysis}, the small set of key tokens is dynamic and input-dependent, which prevents the use of a fixed attention pattern. Therefore, we estimate key positions on the fly using a lightweight pre-computation, as detailed in~\cref{sec:online adjustment}.

\textbf{DINO Top-$K$ Kernel.}
Correspondence heads exhibit clear semantic behavior, attending to relevant patches at corresponding spatial locations across frames.
We observe that DINO patch features already capture such semantic information and can serve as effective correspondence estimators, as visualized in the appendix. Therefore, we pre-compute the cosine similarity between each patch token and all other patch tokens using the DINO features, and select the Top-$K$ most similar tokens from each frame as the keys and values for attention. Empirically, this strategy provides strong coverage of the full attention distribution with a relatively small $K$.

\textbf{Uniform Sampling Kernel.}
For scanning heads, attention patterns are diverse and broadly distributed across tokens. Unlike other head types that focus on a small subset of keys, scanning heads aggregate information from a wide range of spatial locations without clear structural regularity, making pattern-specific key selection prone to missing important interactions. Therefore, following prior work~\cite{shen2025fastvggt,sun2025avggt}, we adopt a uniform sampling kernel that selects key positions with a fixed stride. This strategy provides an effective approximation by evenly covering the token space. In practice, it performs well at relatively high sampling ratios (\ie, retaining 25\%–50\% of keys), but its performance degrades rapidly as the ratio decreases.

\subsection{Offline Head Profiling}
\label{sec:offline profile}

With tailored sparse kernels for each attention type, the remaining challenge is how to assign each attention head to the appropriate category to fully exploit sparsity while maintaining performance.
To this end, we propose a simple yet effective sparsity search strategy based on progressive local substitution. Specifically, the search starts from a baseline configuration $c_{\text{base}}$ that uses a uniform sampling kernel with sparsity ratio $\sigma$. For each head $h$, we then define a candidate set of configurations $\mathcal{C}_{h} = \{c_1, c_2, \dots, c_S\}$ that includes alternating sparse attention kernels (\eg, DINO Top-$K$ and Query-Probe Top-$K$) under different settings (\eg, $K$). All configurations in the candidate set are designed to have \emph{equal or higher sparsity} than the baseline uniform sampling kernel.
Next, we compute the exact full-attention output $\mathbf{O}_{\text{full}}^{(h)} \in \mathbb{R}^{N\times D}$ and the sparse attention output $\mathbf{O}_{\text{sparse}}^{(h,c)} \in \mathbb{R}^{N\times D}$ under configuration $c$, and define the approximation cost using Normalized Mean Squared Error (NMSE):
\begin{equation}
E(c) = \mathbb{E}_{\mathcal{D}_{\text{calib}}}
\left[
\frac{\|\mathbf{O}_{\text{full}}^{(h)} - \mathbf{O}_{\text{sparse}}^{(h,c)}\|_2^2}
{\|\mathbf{O}_{\text{full}}^{(h)}\|_2^2}
\right],
\end{equation}
where $\mathcal{D}_{\text{calib}}$ is a small calibration dataset used for profiling.

We then evaluate candidate sparse configurations using a progressive substitution strategy. 
Starting from the baseline configuration $c_{\text{base}}$, we iteratively update the current best configuration whenever a candidate achieves lower approximation error. 
Formally, the selected configuration minimizes the approximation error under the constraint that its computational complexity does not exceed that of the baseline:
\begin{equation}
c^* =
\underset{c \in \mathcal{C}_{h} \cup \{c_{\text{base}}\}}
{\arg\min}
E(c)
\quad
\text{s.t.}
\quad
\Omega(c) \le \Omega(c_{\text{base}}),
\end{equation}
where $\Omega(c)$ denotes the computational cost of configuration $c$ (measured by floating-point operations).
The progressive replacement produces a heterogeneous head-wise sparsity configuration that preserves the accuracy of the baseline while achieving equal or lower inference cost. Empirically, this simple greedy search significantly increases sparsity while incurring no accuracy loss.

\subsection{Online Pattern Adaptation}
\label{sec:online adjustment}

In offline profiling stage, we determine the appropriate kernel and configuration for each attention head. Due to the dynamic nature of Query-probe and DINO Top-$K$ heads, we further introduce an online adaptation mechanism to refine the attention patterns for each input sequence during inference.

\textbf{Query-Probe Top-$K$ Kernel.}
One intuitive approach is to compute the full pre-softmax attention matrix, average the scores across queries, and select the Top-$K$ keys. However, this requires $\mathcal{O}(N^2)$ attention computation, eliminating any potential speedup. Instead, we show that selecting keys based on the column-wise average of the full attention scores admits an equivalent yet more efficient $\mathcal{O}(N)$ formulation.
Let $Q, K \in \mathbb{R}^{N \times D}$ denote the query and key matrices, and 
$S = QK^\top / \sqrt{D}$ the pre-softmax attention scores. 
By the linearity of inner products, the column-wise average of $S$ across queries can be written as the similarity between each key and the mean query 
$\bar{q} = \frac{1}{N} \sum_{i=1}^{N} q_i$. 
Therefore, selecting the Top-$K$ columns according to the averaged attention scores is equivalent to computing a single probe vector $\bar{q} K^\top / \sqrt{D}$ and ranking keys accordingly.
This preserves the exact ranking induced by column-wise averaging of the full attention scores while incurring only negligible overhead.

\textbf{DINO Top-$K$ Kernel.}
For each input sequence, we extract the output from the last DINO block and compute the global cosine similarity between all patches, storing the indices of the per-frame Top-$K$ matched patches. Although this introduces an $\mathcal{O}(N^2)$ computation, the indices are cached and reused by all subsequent DINO Top-$K$ kernels, thereby amortizing the computational cost.
We explore token-wise and block-wise implementations, which compute exact token-level similarities and block-level similarities averaged over multiple tokens, respectively, offering trade-offs between efficiency and retrieval accuracy.

\section{Experiments}
\label{sec:experiments}

\subsection{Experimental Setup}

\subsubsection{Evaluation benchmarks and metrics.} 
We evaluate F3R transformers for camera pose estimation and point cloud reconstruction under both sparse and dense settings. For sparse evaluation, we uniformly sample 128 frames from each scene (or use all frames if fewer are available) in the following datasets: Co3D-v2~\cite{reizenstein2021common}, RealEstate10K~\cite{zhou2018stereo}, 7Scenes~\cite{shotton2013scene}, HiRoom~\cite{lin2025depth}, ETH3D~\cite{schops2017multi}, ScanNet++~\cite{yeshwanth2023scannet++}, and DTU~\cite{aanaes2016large}. 
For dense evaluation, we uniformly sample 300 - 700 frames from each scene in 7Scenes~\cite{shotton2013scene} and ScanNet~\cite{dai2017scannet}.
We follow the DA3-bench~\cite{lin2025depth} evaluation protocol for both tasks across all datasets, except for ScanNet, where we adopt the evaluation protocol from FastVGGT~\cite{shen2025fastvggt}.

For camera pose evaluation, we report Area Under the Curve (AUC) following~\cite{wang2025vggt,lin2025depth}, where accuracy is defined as the minimum of relative rotation accuracy (RRA) and relative translation accuracy (RTA). Results at $3^\circ$ and $30^\circ$ are reported. We additionally report Absolute Trajectory Error (ATE), Absolute Rotation Error (ARE), and Relative Pose Error (RPE).
For point cloud reconstruction, we report Accuracy, Completeness, Chamfer Distance (CD), and F1-score following~\cite{lin2025depth}.

\subsubsection{Baselines.}
We evaluate on three representative large-scale F3R transformers: VGGT~\cite{wang2025vggt}, $\pi^3$~\cite{wang2025pi}, and Depth Anything 3 (DA3-Giant)~\cite{lin2025depth}, which contain 24, 18, and 14 global attention blocks, respectively.
We compare against state-of-the-art token compression approaches designed for F3R models, including FastVGGT~\cite{shen2025fastvggt}, Co-Me~\cite{chen2025co}, as well as sparse attention methods SparseVGGT~\cite{wang2025faster} and AVGGT~\cite{sun2025avggt}~\footnote{Since no public code is available, we re-implement the method based on the paper.}.
For VGGT, we adopt the memory-efficient implementation from FastVGGT~\cite{shen2025fastvggt}. The merging ratios are set to 0.9 and 0.5 for FastVGGT and Co-Me, respectively, following the official implementations. For SparseVGGT~\cite{wang2025faster}, we adopt a block sparsity ratio of 0.75, while for AVGGT we use a subsampling factor of 4. Except for Co-Me, all compared methods focus on optimizing global attention while leaving the rest of the model unchanged.

\subsubsection{Implementation details.}
The proposed framework is implemented in PyTorch~\cite{ansel2024pytorch} with custom Triton kernels~\cite{tillet2019triton}.
For offline head profiling, we sample 10 training sequences from ETH3D~\cite{schops2017multi} for calibration, spanning indoor and outdoor scenes. We use a global sparsity ratio of $\sigma=0.85$ as the search baseline.
The profiled configurations for each model are provided in the appendix.
During online inference, we set $K$ between 16 and 128 per frame for the DINO Top-$K$ kernel depending on the scene, balancing efficiency and reconstruction accuracy.
All experiments are conducted on a single NVIDIA H100 GPU (80\,GiB VRAM).

\begin{table*}[t]
\centering
\caption{Comparisons of \textbf{camera pose estimation} accuracy under \textbf{sparse settings} (10 - 128 images). We report both AUC@3 ($\uparrow$) and AUC@30 ($\uparrow$) metrics. Results are in percentages (\%). Full-attention baselines are marked in \colorbox{gray!10}{gray}.
}
\label{tab:main cmp sparse pose}
\resizebox{\textwidth}{!}{
\begin{tabular}{@{}l*{14}{c}@{}}
\toprule
\multirow{2}{*}{\textbf{Methods}} & 
\multicolumn{2}{c}{\textbf{Co3D-v2}\cite{reizenstein2021common}} & 
\multicolumn{2}{c}{\textbf{Re10K}\cite{zhou2018stereo}} & 
\multicolumn{2}{c}{\textbf{7Scenes}\cite{shotton2013scene}} & 
\multicolumn{2}{c}{\textbf{ETH3D}\cite{schops2017multi}} & 
\multicolumn{2}{c@{}}{\textbf{ScanNet++}\cite{yeshwanth2023scannet++}} &
\multicolumn{2}{c}{\textbf{HiRoom}\cite{lin2025depth}} & 
\multicolumn{2}{c}{\textbf{DTU}\cite{aanaes2016large}} \\
\cmidrule(lr){2-3} \cmidrule(lr){4-5} \cmidrule(lr){6-7} \cmidrule(lr){8-9} \cmidrule(lr){10-11} \cmidrule(lr){12-13} \cmidrule(l){14-15}
& AUC@30 & AUC@3 & AUC@30 & AUC@3 & AUC@30 & AUC@3 & AUC@30 & AUC@3 & AUC@30 & AUC@3 & AUC@30 & AUC@3 & AUC@30 & AUC@3 \\
\midrule
\rowcolor{gray!10} VGGT\cite{wang2025vggt} & 90.34 & 58.26 & 80.11 & 23.99 & 84.64 & 23.66 & 81.68 & 29.42 & 94.78 & 61.70 & 88.07 & 48.05 & 85.26 & 79.66 \\
FastVGGT\cite{shen2025fastvggt} & 84.84 & 34.73 & 76.99 & 18.39 & 83.01 & 18.67 & \textbf{76.05} & 19.83 & 90.15 & 39.81 & 75.36 & 25.23 & 81.77 & 52.95 \\
SparseVGGT\cite{wang2025faster} & 82.83 & 18.59 & 69.80 & 13.76 & 81.17 & 14.72 & 65.25 & 12.29 & 88.12 & 21.44 & 68.45 & 23.16 & 82.05 & 50.51 \\
Co-Me\cite{chen2025co} & 86.15 & 40.02 & 75.25 & 17.64 & 80.70 & 16.19 & 70.87 & 18.25 & 88.76 & 33.84 & 76.38 & 28.33 & 84.10 & 72.31 \\
AVGGT\cite{sun2025avggt} & 87.78 & 47.10 & 78.22 & 20.85 & \textbf{84.51} & 22.25 & 72.94 & 19.84 & \textbf{92.31} & \textbf{51.57} & 80.54 & 34.22 & \textbf{85.20} & \textbf{79.60} \\
SAF3R (Ours) & \textbf{89.19} & \textbf{52.68} & \textbf{78.35} & \textbf{23.04} & 84.49 & \textbf{23.57} & 75.65 & \textbf{22.49} & 91.33 & 47.65 & 73.93 & \textbf{37.51} & 84.82 & 77.71 \\

\midrule
\rowcolor{gray!10} $\pi^3$\cite{wang2025pi} & 90.91 & 60.83 & 90.56 & 57.03 & 86.15 & 24.09 & 85.57 & 32.70 & 93.83 & 57.05 & 94.47 & 64.73 & 94.85 & 62.95 \\
Fast$\pi^3$\cite{shen2025fastvggt} & 87.56 & 40.49 & 88.40 & 47.20 & 85.30 & 23.80 & 79.32 & 24.20 & 91.46 & 44.28 & 87.24 & 37.92 & 93.03 & 54.24 \\
Sparse$\pi^3$\cite{wang2025faster} & 82.14 & 15.14 & 81.52 & 30.48 & 81.15 & 12.25 & 75.36 & 14.37 & 84.89 & 16.14 & 82.70 & 28.57 & 92.90 & 46.30 \\
A$\pi^3$\cite{sun2025avggt} & 89.27 & 50.69 & 89.06 & 52.11 & \textbf{86.07} & \textbf{24.22} & 78.03 & 22.74 & 92.68 & 50.42 & 90.40 & 46.92 & 93.45 & 55.30 \\
SAF3R (Ours) & \textbf{90.14} & \textbf{53.98} & \textbf{89.69} & \textbf{52.44} & 85.85 & 23.48 & \textbf{83.03} & \textbf{24.30} & \textbf{93.48} & \textbf{50.76} & \textbf{92.13} & \textbf{51.96} & \textbf{94.79} & \textbf{61.46} \\

\midrule
\rowcolor{gray!10} DA3\cite{lin2025depth} & 90.58 & 53.28 & 91.10 & 57.91 & 86.71 & 28.59 & 91.40 & 48.37 & 98.17 & 84.85 & 95.92 & 80.20 & 99.38 & 93.99 \\
FastDA3\cite{shen2025fastvggt} & 87.31 & 43.42 & 88.97 & 52.16 & 85.90 & 28.53 & \textbf{88.53} & \textbf{37.89} & \textbf{94.58} & \textbf{76.36} & 92.93 & 63.23 & 98.82 & 88.69 \\
ADA3\cite{sun2025avggt} & 84.44 & 39.72 & 83.95 & 37.44 & 85.15 & 25.05 & 64.13 & 16.60 & 71.26 & 30.84 & 85.36 & 49.32 & 97.36 & 77.91 \\
SAF3R (Ours) & \textbf{89.73} & \textbf{50.42} & \textbf{89.94} & \textbf{52.92} & \textbf{86.80} & \textbf{28.84} & 87.85 & 35.39 & 92.84 & 62.79 & \textbf{94.68} & \textbf{79.15} & \textbf{98.99} & \textbf{90.39} \\
\bottomrule
\end{tabular}
}
\end{table*}

\begin{table*}[t]
\centering
\caption{
Comparison of \textbf{point cloud reconstruction} performance under \textbf{sparse settings} (10 - 128 images). We report Accuracy (Acc.), Completeness (Comp.), Chamfer Distance (CD), and F1-score (F1). Full-attention baselines are marked in \colorbox{gray!10}{gray}.}
\label{tab:main cmp sparse pcd}
\setlength{\tabcolsep}{3pt}
\resizebox{\textwidth}{!}{
\begin{tabular}{@{}l*{16}{c}@{}}
\toprule
\multirow{2}{*}{\textbf{Methods}} & 
\multicolumn{4}{c}{\textbf{7Scenes}\cite{shotton2013scene}} & 
\multicolumn{4}{c}{\textbf{ETH3D}\cite{schops2017multi}} & 
\multicolumn{4}{c}{\textbf{ScanNet++}\cite{yeshwanth2023scannet++}} &
\multicolumn{4}{c@{}}{\textbf{HiRoom}\cite{lin2025depth}} \\
\cmidrule(lr){2-5} \cmidrule(lr){6-9} \cmidrule(lr){10-13} \cmidrule(l){14-17}
& Acc.$^\downarrow$ & Comp.$^\downarrow$ & CD$^\downarrow$ & F1$^\uparrow$ & Acc.$^\downarrow$ & Comp.$^\downarrow$ & CD$^\downarrow$ & F1$^\uparrow$ & Acc.$^\downarrow$ & Comp.$^\downarrow$ & CD$^\downarrow$ & F1$^\uparrow$ & Acc.$^\downarrow$ & Comp.$^\downarrow$ & CD$^\downarrow$ & F1$^\uparrow$ \\
\midrule
\rowcolor{gray!10} VGGT\cite{wang2025vggt} & 0.155 & 0.125 & 0.140 & 0.480 & 0.306 & 0.831 & 0.568 & 0.634 & 0.045 & 0.092 & 0.069 & 0.692 & 0.107 & 0.092 & 0.100 & 0.554 \\
FastVGGT\cite{shen2025fastvggt} & \textbf{0.127} & 0.134 & \textbf{0.130} & \textbf{0.478} & \textbf{0.355} & \textbf{0.926} & \textbf{0.640} & 0.519 & \textbf{0.055} & 0.109 & \textbf{0.082} & 0.568 & 0.181 & 0.161 & 0.171 & 0.429 \\
SparseVGGT\cite{wang2025faster} & 0.156 & 0.161 & 0.159 & 0.371 & 0.426 & 1.779 & 1.102 & 0.460 & 0.078 & 0.120 & 0.099 & 0.447 & 0.246 & 0.203 & 0.225 & 0.279 \\
Co-Me\cite{chen2025co} & 0.121 & 0.158 & 0.140 & 0.455 & 0.557 & 1.221 & 0.889 & 0.548 & 0.094 & 0.114 & 0.104 & 0.486 & 0.210 & 0.134 & 0.172 & 0.406 \\
AVGGT\cite{sun2025avggt} & 0.152 & 0.127 & 0.139 & 0.447 & 0.414 & 1.301 & 0.858 & 0.486 & 0.067 & 0.108 & 0.088 & \textbf{0.571} & \textbf{0.156} & \textbf{0.123} & \textbf{0.140} & \textbf{0.467} \\
SAF3R (Ours) & 0.144 & \textbf{0.127} & 0.136 & 0.473 & 0.406 & 1.048 & 0.727 & \textbf{0.576} & 0.066 & \textbf{0.108} & 0.087 & 0.562 & 0.257 & 0.184 & 0.220 & 0.459 \\

\midrule
\rowcolor{gray!10} $\pi^3$\cite{wang2025pi} & 0.134 & 0.098 & 0.116 & 0.430 & 0.270 & 0.762 & 0.516 & 0.695 & 0.047 & 0.092 & 0.070 & 0.649 & 0.053 & 0.043 & 0.048 & 0.737 \\
Fast$\pi^3$\cite{shen2025fastvggt} & 0.141 & \textbf{0.094} & 0.117 & \textbf{0.473} & 0.290 & 0.761 & 0.525 & 0.621 & 0.063 & 0.105 & 0.084 & 0.519 & 0.127 & 0.099 & 0.113 & 0.352 \\
Sparse$\pi^3$\cite{wang2025faster} & 0.163 & 0.113 & 0.138 & 0.332 & 0.811 & 0.879 & 0.845 & 0.435 & 0.166 & 0.164 & 0.165 & 0.231 & 0.234 & 0.145 & 0.190 & 0.203 \\
A$\pi^3$\cite{sun2025avggt} & \textbf{0.132} & 0.099 & \textbf{0.115} & 0.433 & 0.292 & 0.826 & 0.559 & 0.665 & 0.054 & 0.097 & 0.075 & 0.569 & 0.099 & \textbf{0.063} & \textbf{0.081} & \textbf{0.561} \\
SAF3R (Ours) & 0.150 & 0.101 & 0.125 & 0.424 & \textbf{0.274} & \textbf{0.760} & \textbf{0.517} & \textbf{0.667} & \textbf{0.051} & \textbf{0.095} & \textbf{0.073} & \textbf{0.598} & \textbf{0.092} & 0.074 & 0.083 & 0.501 \\

\midrule
\rowcolor{gray!10} DA3\cite{lin2025depth} & 0.119 & 0.128 & 0.124 & 0.524 & 0.237 & 0.667 & 0.452 & 0.793 & 0.035 & 0.080 & 0.057 & 0.796 & 0.048 & 0.044 & 0.046 & 0.859 \\
FastDA3\cite{shen2025fastvggt} & 0.123 & 0.145 & 0.134 & 0.529 & 0.290 & 0.711 & 0.501 & \textbf{0.734} & \textbf{0.034} & \textbf{0.082} & \textbf{0.058} & \textbf{0.782} & 0.060 & 0.060 & 0.060 & 0.762 \\
ADA3\cite{sun2025avggt} & \textbf{0.120} & 0.144 & \textbf{0.132} & \textbf{0.533} & 0.792 & 1.727 & 1.260 & 0.420 & 0.172 & 0.193 & 0.182 & 0.467 & 0.163 & 0.097 & 0.130 & 0.631 \\
SAF3R (Ours) & \textbf{0.120} & 0.150 & 0.134 & 0.514 & \textbf{0.281} & \textbf{0.666} & \textbf{0.474} & 0.710 & 0.057 & 0.090 & 0.074 & 0.673 & \textbf{0.053} & \textbf{0.048} & \textbf{0.051} & \textbf{0.842} \\
\bottomrule
\end{tabular}
}
\end{table*}

\begin{table*}[t]
\centering
\caption{Comparisons of \textbf{camera pose and point cloud estimation} on ScanNet-50~\cite{dai2017scannet} dataset under \textbf{dense settings}. The average speedup (Spd.) measured relative to the baseline is reported.  Full-attention baselines are marked in \colorbox{gray!10}{gray}.
}
\label{tab:main cmp dense}
\setlength{\tabcolsep}{2.5pt}
\resizebox{\textwidth}{!}{
\begin{tabular}{@{} l *{15}{c} @{}}
\toprule
\multirow{2}{*}{\textbf{Method}} & 
\multicolumn{5}{c}{\textbf{300 Images}} & 
\multicolumn{5}{c}{\textbf{500 Images}} & 
\multicolumn{5}{c@{}}{\textbf{700 Images}} \\
\cmidrule(lr){2-6} \cmidrule(lr){7-11} \cmidrule(l){12-16}
& ATE$^\downarrow$ & ARE$^\downarrow$ & RPE-T/R$^\downarrow$ & CD$^\downarrow$ & Spd.$^\uparrow$ & ATE$^\downarrow$ & ARE$^\downarrow$ & RPE-T/R$^\downarrow$ & CD$^\downarrow$ & Spd.$^\uparrow$ & ATE$^\downarrow$ & ARE$^\downarrow$ & RPE-T/R$^\downarrow$ & CD$^\downarrow$ & Spd.$^\uparrow$ \\
\midrule
\rowcolor{gray!10} VGGT\cite{wang2025vggt} & 0.14 & 3.64 & 0.03/0.74 & 0.45 & 1.0$\times$ & 0.15 & 4.14 & 0.04/0.98 & 0.47 & 1.0$\times$ & 0.19 & 4.63 & 0.04/0.97 & 0.47 & 1.0$\times$ \\
FastVGGT\cite{shen2025fastvggt} & 0.13 & 3.48 & 0.03/0.77 & 0.44 & 2.8$\times$ & 0.13 & 3.54 & 0.03/0.62 & 0.46 & 3.2$\times$ & 0.14 & 3.72 & 0.03/0.73 & 0.47 & 3.4$\times$ \\
SparseVGGT\cite{wang2025faster} & 0.16 & 4.51 & 0.04/0.96 & 0.48 & 2.5$\times$ & 0.18 & 5.49 & 0.04/1.30 & 0.48 & 2.7$\times$ & 0.21 & 5.85 & 0.06/1.65 & 0.48 & 2.8$\times$ \\
Co-Me\cite{chen2025co} & 0.16 & 3.96 & 0.05/1.24 & 0.45 & 2.6$\times$ & 0.15 & 4.02 & 0.05/1.19 & 0.45  & 3.5$\times$ & 0.18 & 4.58 & 0.06/1.20 & 0.49 & 4.0$\times$ \\
AVGGT\cite{sun2025avggt} & 0.15 & 3.95 & 0.04/0.97 & 0.44 & 3.1$\times$ & 0.16 & 4.31 & 0.04/1.11 & 0.45 & 4.3$\times$ & 0.20 & 5.37 & 0.05/1.60 & 0.46 & 4.6$\times$ \\
SAF3R (Ours) & 0.15	& 3.97 & 0.04/0.85 & 0.45 & 3.2$\times$ & 0.15 & 4.13 & 0.03/0.91 & 0.45 & 4.6$\times$ & 0.17 & 4.47 & 0.04/1.03 & 0.46 & 4.7$\times$ \\

\midrule
\rowcolor{gray!10} $\pi^3$\cite{wang2025pi} & 0.08 & 2.16 & 0.02/0.52 & 0.59 & 1.0$\times$ & 0.08 & 2.17 & 0.01/0.43 & 0.59 & 1.0$\times$ & 0.08 & 2.15 & 0.01/0.37 & 0.59 & 1.0$\times$ \\
Fast$\pi^3$\cite{shen2025fastvggt} & 0.09 & 2.37 & 0.02/0.55 & 0.59 & 2.0$\times$ & 0.09 & 2.35 & 0.02/0.41 & 0.59 & 2.2$\times$ & 0.09 & 2.39 & 0.02/0.45 & 0.60 & 2.3$\times$ \\
Sparse$\pi^3$\cite{wang2025faster} & 0.14 & 4.99 & 0.02/0.61 & 0.58 & 2.7$\times$ & 0.14 & 4.99 & 0.02/0.49 & 0.60 & 2.9$\times$ & 0.14 & 4.98 & 0.02/0.43 & 0.59 & 3.1$\times$ \\
A$\pi^3$\cite{sun2025avggt} & 0.10 & 2.39 & 0.02/0.55 & 0.59 & 4.9$\times$ & 0.09 & 2.36 & 0.02/0.46 & 0.59 & 5.9$\times$ & 0.09 & 2.37 & 0.02/0.40 & 0.60 & 6.7$\times$ \\
SAF3R (Ours) & 0.09 & 2.36 & 0.02/0.54 & 0.58 & 5.3$\times$ & 0.09 & 2.40 & 0.02/0.44 & 0.58 & 6.5$\times$ & 0.09 & 2.44 & 0.01/0.39 & 0.59 & 7.3$\times$ \\

\midrule
\rowcolor{gray!10} DA3\cite{lin2025depth} & 0.13 & 3.68 & 0.02/0.82 & 0.40 & 1.0$\times$ & 0.13 & 3.73 & 0.02/0.77 & 0.40 & 1.0$\times$ & OOM & OOM & OOM & OOM & OOM \\
FastDA3\cite{shen2025fastvggt} & 0.13 & 3.77 & 0.02/1.00 & 0.40 & 2.7$\times$ & 0.13 & 3.94 & 0.02/1.00 & 0.40 & 3.1$\times$ & OOM & OOM & OOM & OOM & OOM \\
ADA3\cite{sun2025avggt} & 0.35 & 12.43 & 0.11/7.34 & 0.43 & 3.2$\times$ & 0.43 & 13.42 & 0.10/7.67 & 0.45 & 3.8$\times$ & OOM & OOM & OOM & OOM & OOM \\
SAF3R (Ours) & 0.15	& 5.19 & 0.03/1.40 & 0.40 & 3.3$\times$ & 0.16 & 5.71 & 0.04/1.89 & 0.43 & 3.8$\times$ & OOM & OOM & OOM & OOM & OOM \\

\bottomrule
\end{tabular}
}
\end{table*}

\subsection{Main Results}

\subsubsection{Camera pose estimation.}
\cref{tab:main cmp sparse pose} presents the comparison of camera pose estimation performance of F3R models under sparse settings. As shown, our method outperforms existing efficient F3R models on most datasets. Notably, it preserves camera pose accuracy even under very strict error tolerances, resulting in only negligible degradation on the challenging AUC@3 metric, while other methods suffer noticeable performance drops.
As shown in~\cref{tab:main cmp dense}, our method shows a slight performance drop under dense settings on the ScanNet dataset, particularly in camera rotation estimation, while still achieving relatively low error compared to the baseline models.

\subsubsection{Point cloud reconstruction.}
We compare the point cloud reconstruction performance of F3R models under sparse and dense settings in~\cref{tab:main cmp sparse pcd} and~\cref{tab:main cmp dense}, respectively. As shown, FastVGGT achieves strong performance on several datasets, even outperforming the baseline models in some cases. However, it relies on dense anchor sampling and therefore achieves relatively limited speedup compared to the baseline. While our method may underperform FastVGGT on certain datasets, it generally achieves reconstruction quality comparable to the baseline models and remains competitive with other efficient approaches, while delivering substantially higher speedup due to dynamic sparse attention.

\subsubsection{Efficiency benchmarks.}

\begin{figure}[t]
  \centering
  \includegraphics[width=\textwidth]{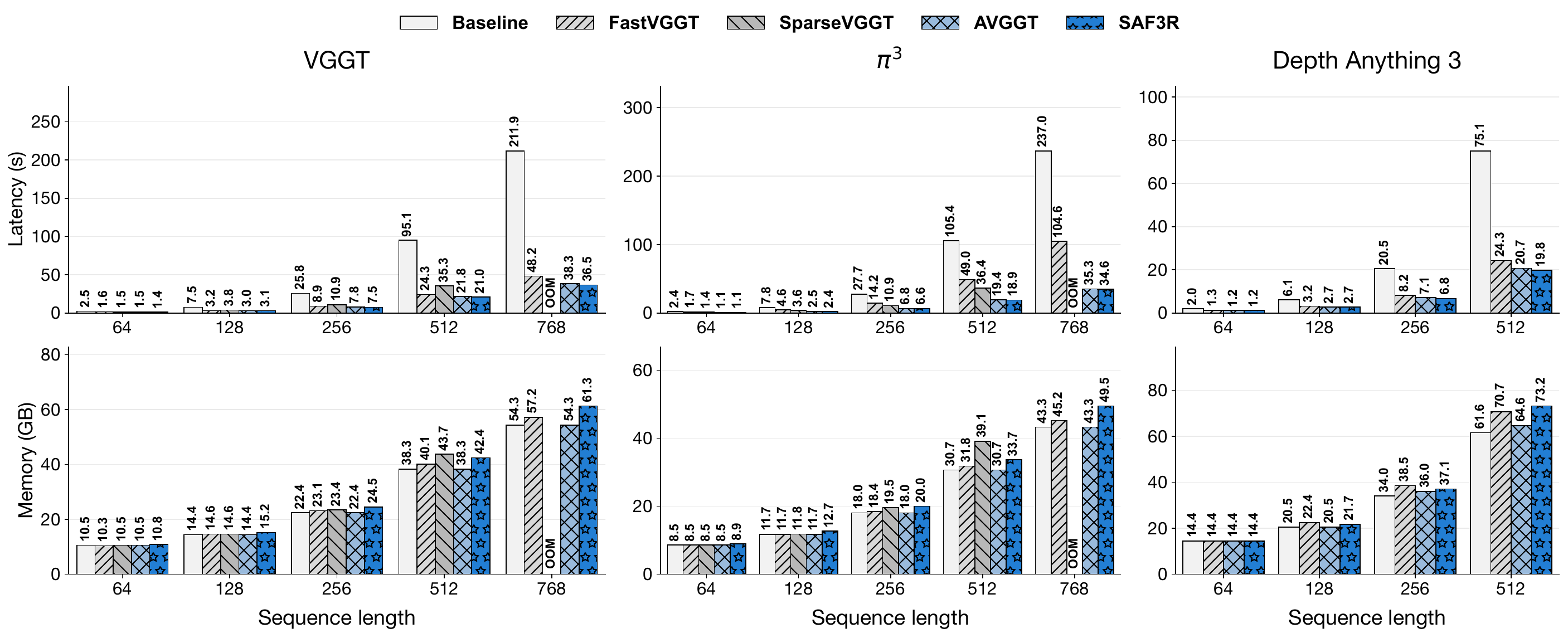}
  \caption{Efficiency results for processing a single scene with different sequence lengths. Latency and memory are measured on scenes with fixed image resolution $480 \times 640$.
  }
  \label{fig:efficiency}
\end{figure}

\cref{fig:efficiency} presents the efficiency benchmarking results of different methods across models when processing a single scene from the 7Scenes dataset.
For short sequences, the bottleneck lies in other components of the model, such as feed-forward networks, and global attention does not dominate the computation. As a result, all models achieve similar performance.
As the sequence length increases, full-attention baselines become dominated by the quadratic cost of global attention. In contrast, with acceptable memory overhead from computing and caching Top-$K$ indices, our method exploits dynamic sparsity in attention computation and achieves end-to-end speedups of up to 5.8$\times$, 6.8$\times$, and 3.9$\times$ for VGGT, $\pi^3$, and DA3, respectively.

\subsubsection{Visualization of offline profiling results.}
\begin{figure}[t]
  \centering
  \includegraphics[width=\textwidth]{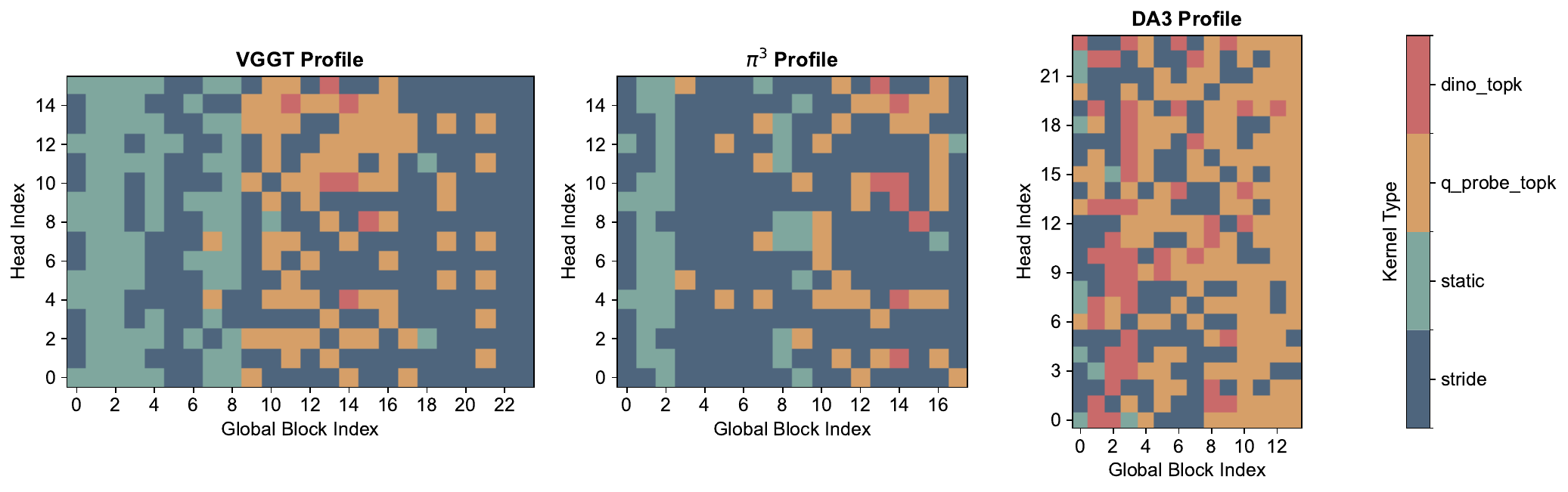}
  \caption{
  Offline head profiling results for different F3R models. After profiling, each global attention head is assigned to one of four predefined kernel types.
  }
  \label{fig:head profile res}
\end{figure}

\cref{fig:head profile res} presents the offline head profiling results. As shown, different models exhibit varying degrees of head-wise heterogeneity, and those with greater heterogeneity benefit more from dynamic sparse attention, highlighting the need for head-wise heterogeneous kernel configurations and the flexibility of our method.

\subsection{Ablation Study}

\begin{table}[t]
\centering
\caption{Effectiveness of the proposed tailored sparse kernels and profiling strategy (DINO Top-$K$=128). Camera pose estimation results (AUC@30 and AUC@3) are reported over five runs with different randomly sampled sets of scene images.}
\label{tab:effect offline profile}
\setlength{\tabcolsep}{5pt}
\scriptsize
\begin{tabular}{@{} c l c *{5}{c} @{}}
\toprule
\textbf{Model} & \textbf{Configs} & \textbf{Sparsity} & \textbf{7Scenes} & \textbf{ETH3D} & \textbf{ScanNet++} & \textbf{HiRoom} & \textbf{DTU} \\
\midrule
\multirow{5}{*}{VGGT}
  & Uniform  & 0.85 & \makecell{83.8$\pm$0.3 \\ 22.6$\pm$0.6} & \makecell{77.2$\pm$0.8 \\ 23.2$\pm$1.4} & \makecell{93.3$\pm$0.2 \\ 53.9$\pm$1.0} & \makecell{82.9$\pm$1.1 \\ 41.2$\pm$1.5} & \makecell{84.5$\pm$0.8 \\ 71.7$\pm$1.1} \\
\cdashline{2-8}
  & +Static  & 0.90 & \makecell{84.4$\pm$0.3 \\ 23.3$\pm$0.5} & \makecell{76.9$\pm$0.8 \\ 23.5$\pm$1.2} & \makecell{93.9$\pm$0.3 \\ 54.3$\pm$0.9} & \makecell{82.2$\pm$1.2 \\ 40.9$\pm$1.1} & \makecell{84.8$\pm$1.0 \\ 72.5$\pm$1.4} \\
\cdashline{2-8}
  & +Dynamic & 0.91 & \makecell{84.8$\pm$0.4 \\ 24.2$\pm$0.6} & \makecell{78.1$\pm$0.9 \\ 24.9$\pm$1.6} & \makecell{93.6$\pm$0.3 \\ 54.8$\pm$0.9} & \makecell{78.8$\pm$1.2 \\ 39.1$\pm$0.9} & \makecell{85.4$\pm$0.8 \\ 74.7$\pm$1.7} \\
\midrule
\multirow{5}{*}{DA3}
  & Uniform  & 0.82 & \makecell{86.8$\pm$0.2 \\ 29.3$\pm$0.4} & \makecell{86.8$\pm$2.1 \\ 36.1$\pm$3.5} & \makecell{96.8$\pm$0.7 \\ 74.2$\pm$2.3} & \makecell{93.6$\pm$0.3 \\ 72.5$\pm$1.1} & \makecell{98.5$\pm$0.5 \\ 88.0$\pm$2.3} \\
\cdashline{2-8}
  & +Static  & 0.85 & \makecell{86.8$\pm$0.3 \\ 29.3$\pm$0.6} & \makecell{88.4$\pm$1.5 \\ 38.4$\pm$3.2} & \makecell{96.8$\pm$0.4 \\ 75.3$\pm$2.2} & \makecell{93.9$\pm$0.7 \\ 71.9$\pm$1.3} & \makecell{99.1$\pm$0.5 \\ 88.5$\pm$2.1} \\
\cdashline{2-8}
  & +Dynamic & 0.87 & \makecell{86.8$\pm$0.2 \\ 29.4$\pm$0.5} & \makecell{89.1$\pm$2.6 \\ 39.3$\pm$3.4} & \makecell{97.5$\pm$0.4 \\ 79.8$\pm$2.6} & \makecell{94.9$\pm$0.9 \\ 81.8$\pm$1.9} & \makecell{98.5$\pm$0.5 \\ 87.2$\pm$2.8} \\
\bottomrule
\end{tabular}
\end{table}

\subsubsection{Effectiveness of tailored sparse kernels and offline search.}
We evaluate the effectiveness of the proposed offline search strategy in \cref{tab:effect offline profile}. 
Starting from a uniform sampling baseline with 85\% sparsity, the search progressively replaces attention kernels with more suitable configurations for each head. 
The resulting model further increases the overall sparsity while preserving model accuracy. 
Specifically, the selected configurations reduce the number of attended keys by an additional 6\% compared to the baseline, while slightly improving reconstruction quality. 
These results demonstrate that the proposed search strategy can effectively exploit heterogeneous attention behaviors across heads to achieve a better sparsity-accuracy trade-off.

\begin{table}[t]
  \centering
  
  \begin{minipage}[t]{0.515\textwidth}
    \centering
    \caption{
      Effectiveness of DINO Top-$K$ selection on 7Scenes dataset. Sparse attention is applied only to heads $\{4, 10, 14\}$ in layer 14 of VGGT.
    }
    \label{tab:ablation dino topk 7scenes}
    \resizebox{\linewidth}{!}{
        \begin{tabular}{@{}lrccc@{}}
          \toprule
          \textbf{Kernel Type} 
          & \textbf{\#Keys} 
          & \textbf{AUC@30}$^\uparrow$ 
          & \textbf{AUC@3}$^\uparrow$ & \textbf{CD}$^\downarrow$ \\
          \midrule
          Full & 100\% & 84.64 & 23.66 & 0.14 \\
          \midrule
          Skip & 0\% & 2.56 & 0.00 & 0.97 \\
          \midrule
          Uniform Stride ($s$=64) & 1.6\% & 82.13 & 19.94 & 0.15 \\
          Uniform Stride ($s$=32) & 3.1\% & 84.45 & 22.64 & 0.14 \\
          \midrule
          DINO Top-$K$ ($K$=2) & 0.2\% & 84.40 & 22.92 & 0.14 \\
          DINO Top-$K$ ($K$=4) & 0.4\% & 84.51 & 23.34 & 0.14 \\
          \bottomrule
        \end{tabular}
    }
  \end{minipage}%
  \hfill
  \begin{minipage}[t]{0.47\textwidth}
    \centering
    \caption{
      Effectiveness of DINO Top-$K$ selection on Co3D-v2 dataset. Sparse attention is applied only to heads $\{4, 10, 14\}$ in layer 14 of VGGT.
    }
    \label{tab:ablation dino topk co3d}
    \resizebox{\linewidth}{!}{
        \begin{tabular}{@{}lrccc@{}}
          \toprule
          \textbf{Kernel Type} 
          & \textbf{\#Keys} 
          & \textbf{AUC@30}$^\uparrow$ 
          & \textbf{AUC@3}$^\uparrow$ \\
          \midrule
          Full & 100\% & 90.34 & 58.26  \\
          \midrule
          Skip & 0\% & 19.55 & 0.03 \\
          \midrule
          Uniform Stride ($s$=64) & 1.6\% & 81.54 & 33.06 \\
          Uniform Stride ($s$=32) & 3.1\% & 86.85 & 43.77 \\
          \midrule
          DINO Top-$K$ ($K$=2) & 0.2\% & 87.44 & 45.34 \\
          DINO Top-$K$ ($K$=4) & 0.4\% & 88.05 & 48.05 \\
          \bottomrule
        \end{tabular}
    }
  \end{minipage}
  
\end{table}

\subsubsection{Effectiveness of DINO Top-$K$ for correspondence heads.}
We compare the impact of different sparse attention kernels on the correspondence heads.
\cref{tab:ablation dino topk 7scenes} and \cref{tab:ablation dino topk co3d} present the results of applying different sparse attention kernels to correspondence heads and their impact on model performance. As shown, skipping these heads results in severe performance degradation. While uniform sampling requires a large sampling ratio to cover the correspondence pattern, our DINO Top-$K$ kernel achieves comparable results while attending to significantly fewer keys, resulting in a 16$\times$ increase in sparsity.

\begin{figure}[t]
  \centering
\includegraphics[width=\linewidth]{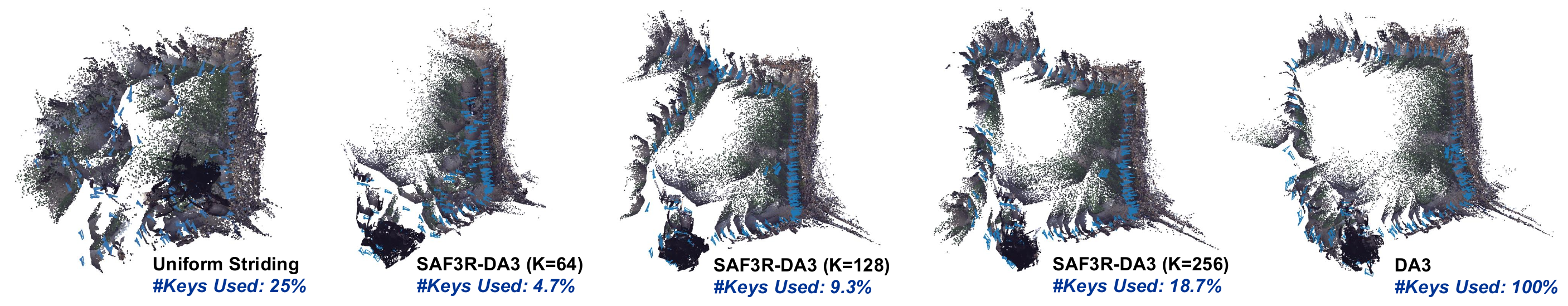}
   \caption{Qualitative comparison across different values of $K$ for the DINO Top-$K$ kernel on the Oxford Spires dataset.}
   \label{fig:da3 oxfordspires}
\end{figure}

\subsubsection{Further Analysis of the DINO Top-$K$ kernel.}
DINO Top-$K$ assumes that DINO features capture patches similar to those used by correspondence heads. This assumption can be violated in practice due to limited feature discriminability (\eg, textureless regions, repeated patterns, illumination changes, and large viewpoint variations). When the assumption fails, correspondence heads may miss critical regions, leading to performance degradation, especially with small $K$.
For example, we observe degradation on VGGT in certain HiRoom scenes, where small $K$ performs worse than uniform sampling. Increasing $K$ effectively mitigates this issue, albeit at the cost of reduced sparsity.
We further evaluate DA3 on a challenging large-scale outdoor scene with loop closures from the Oxford Spires dataset~\cite{tao2025spires}. As shown in \cref{fig:da3 oxfordspires}, increasing $K$
progressively improves trajectory accuracy, whereas uniform
striding performs poorly in this setting.
Overall, the choice of $K$ for DINO kernels reflects a trade-off between
robustness and efficiency, and the optimal value may vary across
scenes.

For future improvements to the DINO Top-$K$ kernel, we identify two promising directions:
(1) more accurate Top-$K$ estimation by deriving indices directly from the first correspondence head instead of DINO features;
(2) combining Top-$K$ with uniform striding to reduce missing regions in challenging scenes.

\section{Conclusions and Limitations}
We present SAF3R, a training-free dynamic sparse attention framework for efficient inference in feed-forward 3D reconstruction (F3R) transformers. Our head-wise analysis reveals that global attention in F3R models exhibits strong heterogeneity and input-dependent sparsity. By exploiting these properties, SAF3R achieves significant end-to-end speedup while preserving reconstruction and pose estimation quality. We hope our findings provide new insights of global attention in F3R transformers and inspire future research on efficient F3R architectures.

\noindent\textbf{Limitations.}
Our method primarily targets key-value sparsity in global attention, and further acceleration could be achieved by exploiting sparsity in other components, query-side redundancy, or integration with token merging. In addition, the DINO Top-$K$ kernel may be less robust in challenging scenes where DINO features do not reliably localize correspondence regions, requiring larger $K$ and thus reducing sparsity.

\section*{Acknowledgements}
This work was supported in part by the National Science Foundation under Grants CNS-2328972, CCF-2324937, CNS-2122320, and CNS-2133267. 
This work was also supported in part by the TetraMem Inc. Research Award.
We acknowledge the computational resources provided by the Pittsburgh Supercomputing Center and the National Center for Supercomputing Applications.

%
%
\bibliographystyle{splncs04}
\bibliography{main}

@String(IJCV  = {Int. J. Comput. Vis.})

@String(CVPR  = {IEEE Conf. Comput. Vis. Pattern Recog.})

@String(ICCV  = {Int. Conf. Comput. Vis.})

@String(ECCV  = {Eur. Conf. Comput. Vis.})

@String(NeurIPS = {Adv. Neural Inform. Process. Syst.})

@String(ICML  = {Int. Conf. Mach. Learn.})

@String(ICLR  = {Int. Conf. Learn. Represent.})

@String(CVPRW = {IEEE Conf. Comput. Vis. Pattern Recog. Worksh.})

@String(IJCV  = {IJCV})

@String(CVPR  = {CVPR})

@String(ICCV  = {ICCV})

@String(ECCV  = {ECCV})

@String(NeurIPS = {NeurIPS})

@String(ICML  = {ICML})

@String(ICLR  = {ICLR})

@String(CVPRW = {CVPRW})

@inproceedings{shotton2013scene,
  title={Scene coordinate regression forests for camera relocalization in RGB-D images},
  author={Shotton, Jamie and Glocker, Ben and Zach, Christopher and Izadi, Shahram and Criminisi, Antonio and Fitzgibbon, Andrew},
  booktitle=CVPR,
  pages={2930--2937},
  year={2013}
}

@inproceedings{schonberger2016structure,
  title={Structure-from-motion revisited},
  author={Schonberger, Johannes L and Frahm, Jan-Michael},
  booktitle=CVPR,
  pages={4104--4113},
  year={2016}
}

@inproceedings{wang2024dust3r,
  title={Dust3r: Geometric 3d vision made easy},
  author={Wang, Shuzhe and Leroy, Vincent and Cabon, Yohann and Chidlovskii, Boris and Revaud, Jerome},
  booktitle=CVPR,
  pages={20697--20709},
  year={2024}
}

@inproceedings{leroy2024grounding,
  title={Grounding image matching in 3d with mast3r},
  author={Leroy, Vincent and Cabon, Yohann and Revaud, J{\'e}r{\^o}me},
  booktitle=ECCV,
  pages={71--91},
  year={2024},
  organization={Springer}
}

@inproceedings{yang2025fast3r,
  title={Fast3r: Towards 3d reconstruction of 1000+ images in one forward pass},
  author={Yang, Jianing and Sax, Alexander and Liang, Kevin J and Henaff, Mikael and Tang, Hao and Cao, Ang and Chai, Joyce and Meier, Franziska and Feiszli, Matt},
  booktitle=CVPR,
  pages={21924--21935},
  year={2025}
}

@inproceedings{wang2024vggsfm,
  title={Vggsfm: Visual geometry grounded deep structure from motion},
  author={Wang, Jianyuan and Karaev, Nikita and Rupprecht, Christian and Novotny, David},
  booktitle=CVPR,
  pages={21686--21697},
  year={2024}
}

@inproceedings{yang2024depth,
  title={Depth anything: Unleashing the power of large-scale unlabeled data},
  author={Yang, Lihe and Kang, Bingyi and Huang, Zilong and Xu, Xiaogang and Feng, Jiashi and Zhao, Hengshuang},
  booktitle=CVPR,
  pages={10371--10381},
  year={2024}
}

@inproceedings{wang2024cut3r,
  title={Continuous 3D Perception Model with Persistent State},
  author={Qianqian Wang and Yifei Zhang and Aleksander Holynski and Alexei A. Efros and Angjoo Kanazawa},
  booktitle=CVPR,
  year={2025},
  pages={10510-10522},
}

@inproceedings{wang2025vggt,
  title={Vggt: Visual geometry grounded transformer},
  author={Wang, Jianyuan and Chen, Minghao and Karaev, Nikita and Vedaldi, Andrea and Rupprecht, Christian and Novotny, David},
  booktitle = CVPR,
  pages={5294--5306},
  year={2025}
}

@article{wang2025pi,
  title={{$\pi^{3}$: Permutation-Equivariant Visual Geometry Learning}},
  author={Wang, Yifan and Zhou, Jianjun and Zhu, Haoyi and Chang, Wenzheng and Zhou, Yang and Li, Zizun and Chen, Junyi and Pang, Jiangmiao and Shen, Chunhua and He, Tong},
  journal={arXiv preprint arXiv:2507.13347},
  year={2025}
}

@article{keetha2025mapanything,
  title={Mapanything: Universal feed-forward metric 3d reconstruction},
  author={Keetha, Nikhil and M{\"u}ller, Norman and Sch{\"o}nberger, Johannes and Porzi, Lorenzo and Zhang, Yuchen and Fischer, Tobias and Knapitsch, Arno and Zauss, Duncan and Weber, Ethan and Antunes, Nelson and others},
  journal={arXiv preprint arXiv:2509.13414},
  year={2025}
}

@article{lin2025depth,
  title={Depth anything 3: Recovering the visual space from any views},
  author={Lin, Haotong and Chen, Sili and Liew, Junhao and Chen, Donny Y and Li, Zhenyu and Shi, Guang and Feng, Jiashi and Kang, Bingyi},
  journal={arXiv preprint arXiv:2511.10647},
  year={2025}
}

@article{shen2025fastvggt,
  title={Fastvggt: Training-free acceleration of visual geometry transformer},
  author={Shen, You and Zhang, Zhipeng and Qu, Yansong and Zheng, Xiawu and Ji, Jiayi and Zhang, Shengchuan and Cao, Liujuan},
  journal={arXiv preprint arXiv:2509.02560},
  year={2025}
}

@article{wang2025faster,
  title={Faster vggt with block-sparse global attention},
  author={Wang, Chung-Shien Brian and Schmidt, Christian and Piekenbrinck, Jens and Leibe, Bastian},
  journal={arXiv preprint arXiv:2509.07120},
  year={2025}
}

@article{chen2025co,
  title={Co-Me: Confidence-Guided Token Merging for Visual Geometric Transformers},
  author={Chen, Yutian and Qiu, Yuheng and Li, Ruogu and Agha, Ali and Omidshafiei, Shayegan and Patrikar, Jay and Scherer, Sebastian},
  journal={arXiv preprint arXiv:2511.14751},
  year={2025}
}

@article{wang2025flashvggt,
  title={FlashVGGT: Efficient and Scalable Visual Geometry Transformers with Compressed Descriptor Attention},
  author={Wang, Zipeng and Xu, Dan},
  journal={arXiv preprint arXiv:2512.01540},
  year={2025}
}

@article{shu2025litevggt,
  title={LiteVGGT: Boosting Vanilla VGGT via Geometry-aware Cached Token Merging},
  author={Shu, Zhijian and Lin, Cheng and Xie, Tao and Yin, Wei and Li, Ben and Pu, Zhiyuan and Li, Weize and Yao, Yao and Cao, Xun and Guo, Xiaoyang and others},
  journal={arXiv preprint arXiv:2512.04939},
  year={2025}
}

@article{wang2025httm,
  title={HTTM: Head-wise Temporal Token Merging for Faster VGGT},
  author={Wang, Weitian and Meiner, Lukas and Shubham, Rai and De La Parra, Cecilia and Kumar, Akash},
  journal={arXiv preprint arXiv:2511.21317},
  year={2025}
}

@article{sun2025avggt,
  title={AVGGT: Rethinking Global Attention for Accelerating VGGT},
  author={Sun, Xianbing and Zhu, Zhikai and Lou, Zhengyu and Yang, Bo and Tang, Jinyang and Zhang, Liqing and Wang, He and Zhang, Jianfu},
  journal={arXiv preprint arXiv:2512.02541},
  year={2025}
}

@inproceedings{schops2017multi,
  title={A multi-view stereo benchmark with high-resolution images and multi-camera videos},
  author={Schops, Thomas and Schonberger, Johannes L and Galliani, Silvano and Sattler, Torsten and Schindler, Konrad and Pollefeys, Marc and Geiger, Andreas},
  booktitle=CVPR,
  pages={3260--3269},
  year={2017}
}

@article{aanaes2016large,
  title={Large-scale data for multiple-view stereopsis},
  author={Aan{\ae}s, Henrik and Jensen, Rasmus Ramsb{\o}l and Vogiatzis, George and Tola, Engin and Dahl, Anders Bjorholm},
  journal=IJCV,
  volume={120},
  number={2},
  pages={153--168},
  year={2016},
  publisher={Springer}
}

@inproceedings{yeshwanth2023scannet++,
  title={Scannet++: A high-fidelity dataset of 3d indoor scenes},
  author={Yeshwanth, Chandan and Liu, Yueh-Cheng and Nie{\ss}ner, Matthias and Dai, Angela},
  booktitle=ICCV,
  pages={12--22},
  year={2023}
}

@inproceedings{dai2017scannet,
  title={Scannet: Richly-annotated 3d reconstructions of indoor scenes},
  author={Dai, Angela and Chang, Angel X and Savva, Manolis and Halber, Maciej and Funkhouser, Thomas and Nie{\ss}ner, Matthias},
  booktitle=CVPR,
  pages={5828--5839},
  year={2017}
}

@inproceedings{reizenstein2021common,
  title={Common objects in 3d: Large-scale learning and evaluation of real-life 3d category reconstruction},
  author={Reizenstein, Jeremy and Shapovalov, Roman and Henzler, Philipp and Sbordone, Luca and Labatut, Patrick and Novotny, David},
  booktitle=ICCV,
  pages={10901--10911},
  year={2021}
}

@article{zhou2018stereo,
  title={Stereo magnification: Learning view synthesis using multiplane images},
  author={Zhou, Tinghui and Tucker, Richard and Flynn, John and Fyffe, Graham and Snavely, Noah},
  journal={arXiv preprint arXiv:1805.09817},
  year={2018}
}

@inproceedings{yuan2025native,
  title={Native sparse attention: Hardware-aligned and natively trainable sparse attention},
  author={Yuan, Jingyang and Gao, Huazuo and Dai, Damai and Luo, Junyu and Zhao, Liang and Zhang, Zhengyan and Xie, Zhenda and Wei, Yuxing and Wang, Lean and Xiao, Zhiping and others},
  booktitle={Proceedings of the 63rd Annual Meeting of the Association for Computational Linguistics (Volume 1: Long Papers)},
  pages={23078--23097},
  year={2025}
}

@article{zhang2025fast,
  title={Fast video generation with sliding tile attention},
  author={Zhang, Peiyuan and Chen, Yongqi and Su, Runlong and Ding, Hangliang and Stoica, Ion and Liu, Zhengzhong and Zhang, Hao},
  journal={arXiv preprint arXiv:2502.04507},
  year={2025}
}

@inproceedings{tillet2019triton,
  title={Triton: an intermediate language and compiler for tiled neural network computations},
  author={Tillet, Philippe and Kung, Hsiang-Tsung and Cox, David},
  booktitle={Proceedings of the 3rd ACM SIGPLAN International Workshop on Machine Learning and Programming Languages},
  pages={10--19},
  year={2019}
}

@inproceedings{zhang2025spargeattn,
  title={Spargeattn: Accurate sparse attention accelerating any model inference},
  author={Zhang, Jintao and Xiang, Chendong and Huang, Haofeng and Wei, Jia and Xi, Haocheng and Zhu, Jun and Chen, Jianfei},
  booktitle=ICML,
  year={2025}
}

@article{xiao2023efficient,
  title={Efficient streaming language models with attention sinks},
  author={Xiao, Guangxuan and Tian, Yuandong and Chen, Beidi and Han, Song and Lewis, Mike},
  journal={arXiv preprint arXiv:2309.17453},
  year={2023}
}

@article{xiao2024duoattention,
  title={Duoattention: Efficient long-context llm inference with retrieval and streaming heads},
  author={Xiao, Guangxuan and Tang, Jiaming and Zuo, Jingwei and Guo, Junxian and Yang, Shang and Tang, Haotian and Fu, Yao and Han, Song},
  journal={arXiv preprint arXiv:2410.10819},
  year={2024}
}

@inproceedings{xu2025xattention,
  title     = {XAttention: Block Sparse Attention with Antidiagonal Scoring},
  author    = {Xu, Ruyi and Xiao, Guangxuan and Huang, Haofeng and Guo, Junxian and Han, Song},
  booktitle = ICML,
  year      = {2025}
}

@article{jiang2024minference,
  title={Minference 1.0: Accelerating pre-filling for long-context llms via dynamic sparse attention},
  author={Jiang, Huiqiang and Li, Yucheng and Zhang, Chengruidong and Wu, Qianhui and Luo, Xufang and Ahn, Surin and Han, Zhenhua and Abdi, Amir H and Li, Dongsheng and Lin, Chin-Yew and others},
  journal=NeurIPS,
  volume={37},
  pages={52481--52515},
  year={2024}
}

@article{lai2025flexprefill,
  title={Flexprefill: A context-aware sparse attention mechanism for efficient long-sequence inference},
  author={Lai, Xunhao and Lu, Jianqiao and Luo, Yao and Ma, Yiyuan and Zhou, Xun},
  journal={arXiv preprint arXiv:2502.20766},
  year={2025}
}

@article{xi2025sparse,
  title={Sparse VideoGen: Accelerating Video Diffusion Transformers with Spatial-Temporal Sparsity},
  author={Xi, Haocheng and Yang, Shuo and Zhao, Yilong and Xu, Chenfeng and Li, Muyang and Li, Xiuyu and Lin, Yujun and Cai, Han and Zhang, Jintao and Li, Dacheng and others},
  journal={arXiv preprint arXiv:2502.01776},
  year={2025}
}

@article{yang2025sparse,
  title={Sparse VideoGen2: Accelerate Video Generation with Sparse Attention via Semantic-Aware Permutation},
  author={Yang, Shuo and Xi, Haocheng and Zhao, Yilong and Li, Muyang and Zhang, Jintao and Cai, Han and Lin, Yujun and Li, Xiuyu and Xu, Chenfeng and Peng, Kelly and others},
  journal={arXiv preprint arXiv:2505.18875},
  year={2025}
}

@article{bolya2022token,
  title={Token merging: Your vit but faster},
  author={Bolya, Daniel and Fu, Cheng-Yang and Dai, Xiaoliang and Zhang, Peizhao and Feichtenhofer, Christoph and Hoffman, Judy},
  journal={arXiv preprint arXiv:2210.09461},
  year={2022}
}

@inproceedings{bolya2023token,
  title={Token merging for fast stable diffusion},
  author={Bolya, Daniel and Hoffman, Judy},
  booktitle=CVPRW,
  pages={4599--4603},
  year={2023}
}

@article{darcet2023vision,
  title={Vision transformers need registers},
  author={Darcet, Timoth{\'e}e and Oquab, Maxime and Mairal, Julien and Bojanowski, Piotr},
  journal={arXiv preprint arXiv:2309.16588},
  year={2023}
}

@inproceedings{caron2021emerging,
  title={Emerging properties in self-supervised vision transformers},
  author={Caron, Mathilde and Touvron, Hugo and Misra, Ishan and J{\'e}gou, Herv{\'e} and Mairal, Julien and Bojanowski, Piotr and Joulin, Armand},
  booktitle=ICCV,
  pages={9650--9660},
  year={2021}
}

@article{oquab2023dinov2,
  title={Dinov2: Learning robust visual features without supervision},
  author={Oquab, Maxime and Darcet, Timoth{\'e}e and Moutakanni, Th{\'e}o and Vo, Huy and Szafraniec, Marc and Khalidov, Vasil and Fernandez, Pierre and Haziza, Daniel and Massa, Francisco and El-Nouby, Alaaeldin and others},
  journal={arXiv preprint arXiv:2304.07193},
  year={2023}
}

@article{mur2017orb,
  title={Orb-slam2: An open-source slam system for monocular, stereo, and rgb-d cameras},
  author={Mur-Artal, Raul and Tard{\'o}s, Juan D},
  journal={IEEE Transactions on Robotics},
  volume={33},
  number={5},
  pages={1255--1262},
  year={2017},
  publisher={IEEE}
}

@article{furukawa2015multi,
  title={Multi-view stereo: A tutorial},
  author={Furukawa, Yasutaka and Hern{\'a}ndez, Carlos},
  journal={Foundations and Trends in Computer Graphics and Vision},
  volume={9},
  number={1-2},
  pages={1--148},
  year={2015},
  publisher={Emerald Publishing Limited}
}

@inproceedings{schonberger2016pixelwise,
  title={Pixelwise view selection for unstructured multi-view stereo},
  author={Sch{\"o}nberger, Johannes L and Zheng, Enliang and Frahm, Jan-Michael and Pollefeys, Marc},
  booktitle=ECCV,
  pages={501--518},
  year={2016},
  organization={Springer}
}

@article{vaswani2017attention,
  title={Attention is all you need},
  author={Vaswani, Ashish and Shazeer, Noam and Parmar, Niki and Uszkoreit, Jakob and Jones, Llion and Gomez, Aidan N and Kaiser, {\L}ukasz and Polosukhin, Illia},
  journal=NeurIPS,
  volume={30},
  year={2017}
}

@article{ren2026speed3r,
    title={Speed3R: Sparse Feed-forward 3D Reconstruction Models},
    author={Ren, Weining and Tan, Xiao and Han, Kai},
    journal={arXiv preprint arXiv:2603.08055},
    year={2026}
}

@inproceedings{ansel2024pytorch,
  title={Pytorch 2: Faster machine learning through dynamic python bytecode transformation and graph compilation},
  author={Ansel, Jason and Yang, Edward and He, Horace and Gimelshein, Natalia and Jain, Animesh and Voznesensky, Michael and Bao, Bin and Bell, Peter and Berard, David and Burovski, Evgeni and others},
  booktitle={Proceedings of the 29th ACM international conference on architectural support for programming languages and operating systems, volume 2},
  pages={929--947},
  year={2024}
}

@article{dosovitskiy2020image,
  title={An image is worth 16x16 words: Transformers for image recognition at scale},
  author={Dosovitskiy, Alexey and Beyer, Lucas and Kolesnikov, Alexander and Weissenborn, Dirk and Zhai, Xiaohua and Unterthiner, Thomas and Dehghani, Mostafa and Minderer, Matthias and Heigold, Georg and Gelly, Sylvain and others},
  journal={arXiv preprint arXiv:2010.11929},
  year={2020}
}

@article{tao2025spires,
  title={The oxford spires dataset: Benchmarking large-scale lidar-visual localisation, reconstruction and radiance field methods},
  author={Tao, Yifu and Mu{\~n}oz-Ba{\~n}{\'o}n, Miguel {\'A}ngel and Zhang, Lintong and Wang, Jiahao and Fu, Lanke Frank Tarimo and Fallon, Maurice},
  journal={The International Journal of Robotics Research},
  volume={45},
  number={6},
  pages={839--857},
  year={2026},
  publisher={SAGE Publications Sage UK: London, England}
}

@inproceedings{fang2026dens3r,
  title={Dens3r: A foundation model for 3d geometry prediction},
  author={Fang, Xianze and Gao, Jingnan and Wang, Zhe and Chen, Zhuo and Ren, Xingyu and Lyu, Jiangjing and Ren, Qiao-Mu and Yang, Zhonglei and Yang, Xiaokang and Yan, Yichao and others},
  booktitle=ICLR,
  volume={2026},
  pages={80604--80634},
  year={2026}
}

@inproceedings{gao2026more,
  title={More: 3d visual geometry reconstruction meets mixture-of-experts},
  author={Gao, Jingnan and Wang, Zhe and Fang, Xianze and Ren, Xingyu and Chen, Zhuo and Liu, Shengqi and Cheng, Yuhao and Lyu, Jiangjing and Yang, Xiaokang and Yan, Yichao},
  booktitle=CVPR,
  pages={14680--14691},
  year={2026}
}

@article{wang2026vggto,
  title={{VGGT}-{$\Omega$}},
  author={Wang, Jianyuan and Chen, Minghao and Zhang, Shangzhan and Karaev, Nikita and Sch{\"o}nberger, Johannes and Labatut, Patrick and Bojanowski, Piotr and Novotny, David and Vedaldi, Andrea and Rupprecht, Christian},
  journal={arXiv preprint arXiv:2605.15195},
  year={2026}
}

@inproceedings{zhuo2026streaming,
  title={Streaming visual geometry transformer},
  author={Zhuo, Dong and Zheng, Wenzhao and Guo, Jiahe and Wu, Yuqi and Zhou, Jie and Lu, Jiwen},
  booktitle=ICLR,
  volume={2026},
  pages={88055--88072},
  year={2026}
}

@inproceedings{chen2026ttt3r,
  title={Ttt3r: 3d reconstruction as test-time training},
  author={Chen, Xingyu and Chen, Yue and Xiu, Yuliang and Geiger, Andreas and Chen, Anpei},
  booktitle=ICLR,
  volume={2026},
  pages={50694--50718},
  year={2026}
}

@inproceedings{lan2026stream3r,
  title={Stream3r: Scalable sequential 3d reconstruction with causal transformer},
  author={Lan, Yushi and Luo, Yihang and Hong, Fangzhou and Zhou, Shangchen and Chen, Honghua and Lyu, Zhaoyang and Dai, Bo and Yang, Shuai and Loy, Chen Change and Pan, Xingang},
  booktitle=ICLR,
  volume={2026},
  pages={42746--42768},
  year={2026}
}

@article{chen2026geometric,
  title={Geometric context transformer for streaming 3d reconstruction},
  author={Chen, Lin-Zhuo and Gao, Jian and Chen, Yihang and Cheng, Ka Leong and Sun, Yipengjing and Hu, Liangxiao and Xue, Nan and Zhu, Xing and Shen, Yujun and Yao, Yao and others},
  journal={arXiv preprint arXiv:2604.14141},
  year={2026}
}

@article{wang2026stac,
  title={STAC: Plug-and-play spatio-temporal aware cache compression for streaming 3D reconstruction},
  author={Wang, Runze and Song, Yuxuan and Cai, Youcheng and Liu, Ligang},
  journal={arXiv preprint arXiv:2603.20284},
  year={2026}
}

@article{wu2026point3r,
  title={Point3r: Streaming 3d reconstruction with explicit spatial pointer memory},
  author={Wu, Yuqi and Zheng, Wenzhao and Zhou, Jie and Lu, Jiwen},
  journal=NeurIPS,
  volume={38},
  pages={69675--69699},
  year={2026}
}

@article{yu2026towards,
  title={Towards Consistent Video Geometry Estimation},
  author={Yu, Zhu and Gao, Jingnan and Zhang, Runmin and Qiu, Lingteng and Zhao, Zhengyi and Peng, Rui and Yan, Yichao and Qiu, Kejie and Zhu, Siyu and Dong, Zilong and others},
  journal={arXiv preprint arXiv:2605.30060},
  year={2026}
}

@article{zhang2026loger,
  title={Loger: Long-context geometric reconstruction with hybrid memory},
  author={Zhang, Junyi and Herrmann, Charles and Hur, Junhwa and Sun, Chen and Yang, Ming-Hsuan and Cole, Forrester and Darrell, Trevor and Sun, Deqing},
  journal={arXiv preprint arXiv:2603.03269},
  year={2026}
}

@inproceedings{jin2026zipmap,
  title={Zipmap: Linear-time stateful 3d reconstruction via test-time training},
  author={Jin, Haian and Wu, Rundi and Zhang, Tianyuan and Gao, Ruiqi and Barron, Jonathan T and Snavely, Noah and Ho{\l}y{\'n}ski, Aleksander},
  booktitle=CVPR,
  pages={21748--21759},
  year={2026}
}

@article{elflein2026vgg,
  title={{VGG-T}$^{3}$: Offline Feed-Forward 3D Reconstruction at Scale},
  author={Elflein, Sven and Li, Ruilong and Agostinho, S{\'e}rgio and Gojcic, Zan and Leal-Taix{\'e}, Laura and Zhou, Qunjie and Osep, Aljosa},
  journal={arXiv preprint arXiv:2602.23361},
  year={2026}
}

@article{xie2026scal3r,
  title={Scal3r: Scalable test-time training for large-scale 3d reconstruction},
  author={Xie, Tao and Yang, Peishan and Jin, Yudong and Cai, Yingfeng and Yin, Wei and Ren, Weiqiang and Zhang, Qian and Hua, Wei and Peng, Sida and Guo, Xiaoyang and others},
  journal={arXiv preprint arXiv:2604.08542},
  year={2026}
}

@article{cheng2026horizonstream,
  title={HorizonStream: Long-Horizon Attention for Streaming 3D Reconstruction},
  author={Cheng, Chong and Tao, Peilin and Yao, Nanjie and Ding, Guanzhi and Chen, Xianda and Du, Yuansen and Guo, Xiaoyang and Yin, Wei and Ren, Weiqiang and Zhang, Qian and others},
  journal={arXiv preprint arXiv:2605.23889},
  year={2026}
}

\clearpage
\appendix
\section*{Appendix}

\section{Additional Results on Attention Analysis}

\begin{figure}[h!]
  \centering
  \includegraphics[width=\textwidth]{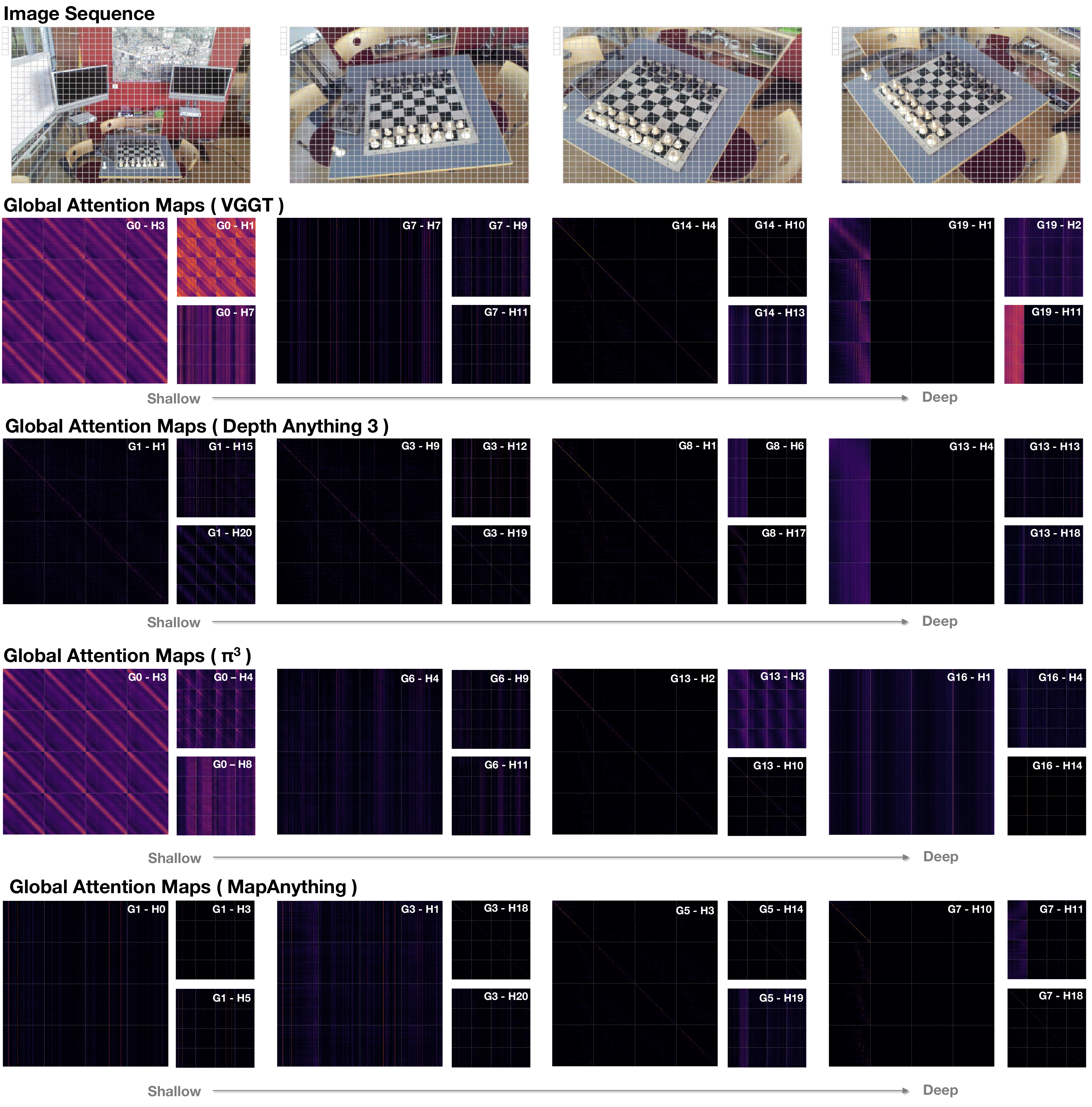}
  \caption{Representative global attention patterns across layers (G) and heads (H) in four F3R transformers on the \textbf{7Scenes Chess} scene. Gray dashed lines in the attention maps separate tokens from different images. Zoom in for best view.
  }
  \label{appendix fig:attn pat vis 1}
\end{figure}

\begin{figure}[h!]
  \centering
  \includegraphics[width=\textwidth]{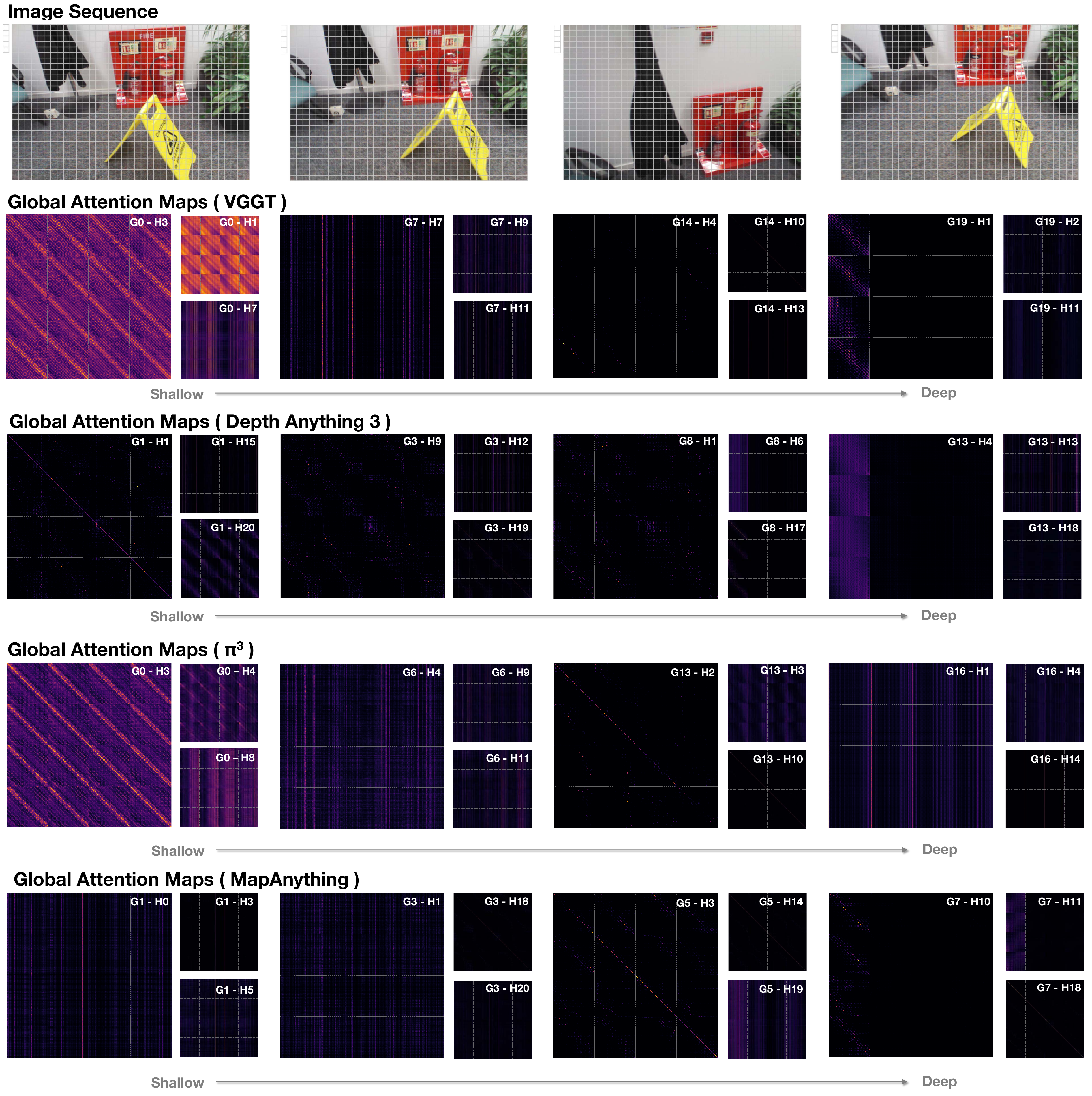}
  \caption{Representative global attention patterns across layers (G) and heads (H) in four F3R transformers on the \textbf{7Scenes Fire} scene. Gray dashed lines in the attention maps separate tokens from different images. Zoom in for best view.
  }
  \label{appendix fig:attn pat vis 2}
\end{figure}

\subsubsection{Full visualization results of four F3R transformers.}
\cref{appendix fig:attn pat vis 1} and \cref{appendix fig:attn pat vis 2} provide the full visualization results on two different scenes for four representative F3R transformers: VGGT~\cite{wang2025vggt}, $\pi^3$~\cite{wang2025pi}, MapAnything~\cite{keetha2025mapanything}, and Depth Anything 3 (DA3)~\cite{lin2025depth}. As shown, these models exhibit similar three-stage attention patterns, consistent with the observations discussed in the main paper. Nevertheless, nuanced differences exist. For example, since the $\pi^3$ model~\cite{wang2025pi} does not rely on a camera reference frame, it does not exhibit the attention pattern where all frames attend to a designated reference frame. In addition, because the MapAnything model~\cite{keetha2025mapanything} does not include positional encoding before the attention operation, no positional attention heads are observed. The heterogeneity of models further demonstrates that a rigid strategy does not work for all models and highlights the need for the proposed heterogeneous head-wise pattern profiling.

\subsubsection{Visual comparison of Top-$K$ selected patches from correspondence heads and those from DINO similarity.}
\begin{figure}[h!]
  \centering
  \includegraphics[width=\textwidth]{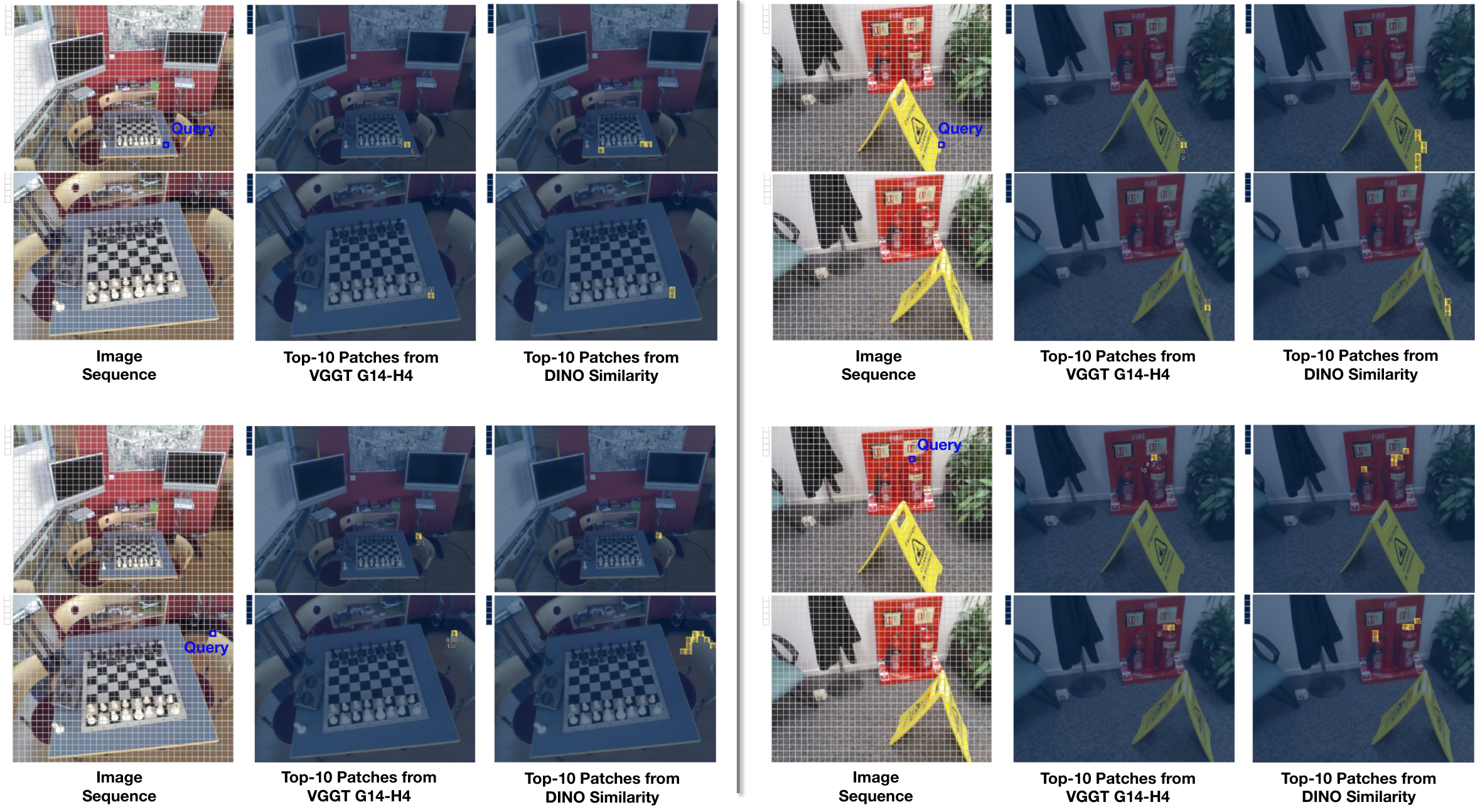}
  \caption{Comparison of the Top-10 selected keys from a \emph{correspondence head} (VGGT-G14-H4) and those selected based on DINO patch similarity. As shown, DINO patch similarity exhibits a high overlap with the patches selected by correspondence heads, especially covering the corresponding patches at the same spatial locations, motivating the use for sparse attention in these heads.
  }
  \label{fig:dino motivation}
\end{figure}

\cref{fig:dino motivation} illustrates the Top-$K$ ($K$=10) keys selected by a correspondence head in VGGT and the Top-$K$ most similar patches computed using cosine similarity of DINO output features. As shown, there is a substantial overlap between the selected patches. This observation motivates us to precompute the Top-$K$ similarity indices using DINO output features for each patch and cache them for subsequent use in correspondence heads, as implemented in our DINO Top-$K$ kernels.

\section{More Experimental Results}

\begin{table*}[h!]
\centering
\caption{Comparisons of {camera pose and point cloud estimation} on 7Scenes~\cite{shotton2013scene} dataset under {dense settings}. Baselines are marked in \colorbox{gray!10}{gray}.}
\label{tab:7scenes cmp dense}
\setlength{\tabcolsep}{4pt}
\resizebox{\textwidth}{!}{
\begin{tabular}{@{} l *{8}{c} @{}}
\toprule
\multirow{2}{*}{{Method}} & 
\multicolumn{4}{c}{{300 Images}} & 
\multicolumn{4}{c}{{500 Images}} \\
\cmidrule(lr){2-5} \cmidrule(l){6-9}
& AUC@30$^\uparrow$ & AUC@3$^\uparrow$ & CD$^\downarrow$ & F1$^\uparrow$ & AUC@30$^\uparrow$ & AUC@3$^\uparrow$ & CD$^\downarrow$ & F1$^\uparrow$ \\
\midrule
\rowcolor{gray!10} VGGT\cite{wang2025vggt} & 83.99 & 22.51 & 0.155 & 0.468 & 83.28 & 21.37 & 0.169 & 0.432 \\
FastVGGT\cite{shen2025fastvggt} & 83.12 & 20.29 & 0.152 & {0.470} & 82.65 & 20.06 & {0.159} & {0.450} \\
SparseVGGT\cite{wang2025faster} & 80.81 & 14.49 & 0.166 & 0.372 & 80.40 & 14.24 & 0.173 & 0.363 \\
AVGGT\cite{sun2025avggt} & 83.72 & 20.49 & 0.154 & 0.463 & 82.50 & 18.76 & 0.173 & 0.413 \\
SAF3R (Ours) & {84.26} & {22.65} & {0.152} & 0.468 & {83.47} & {21.58} & 0.161 & 0.447 \\

\midrule
\rowcolor{gray!10} $\pi^3$\cite{wang2025pi} & 86.02 & 24.20 & 0.126 & 0.416 & 85.95 & 24.04 & 0.130 & 0.416 \\
Fast$\pi^3$\cite{shen2025fastvggt} & 85.15 & 23.33 & {0.132} & {0.443} & 85.57 & 23.39 & 0.137 & {0.454} \\
Sparse$\pi^3$\cite{wang2025faster} & 80.95 & 12.39 & 0.150 & 0.329 & 80.84 & 12.56 & 0.156 & 0.319 \\
A$\pi^3$\cite{sun2025avggt} & 84.90 & 22.92 & 0.138 & 0.426 & 84.67 & 23.12 & 0.138 & 0.438 \\
SAF3R (Ours) & {85.64} & {23.53} & 0.136 & 0.420 & {85.62} & {23.42} & {0.137} & 0.431 \\

\midrule
\rowcolor{gray!10} DA3\cite{lin2025depth} & 86.52 & 28.17 & 0.132 & 0.539 & 86.38 & 27.83 & 0.141 & 0.529 \\
FastDA3\cite{shen2025fastvggt} & 86.38 & 28.43 & {0.147} & {0.529} & 86.16 & 28.01 & {0.149} & 0.525 \\
ADA3\cite{sun2025avggt} & 84.90 & 24.62 & 0.152 & 0.516 & 84.67 & 22.26 & 0.151 & 0.513 \\
SAF3R (Ours) & {86.60} & {28.54} & 0.151 & 0.516 & {86.25} & {28.10} & 0.155 & {0.528} \\

\bottomrule
\end{tabular}
}
\end{table*}

\subsubsection{Evaluation on 7Scenes under dense settings.}
\cref{tab:7scenes cmp dense} provides comparisons of both camera pose estimation and point cloud reconstruction results on the 7Scenes dataset~\cite{shotton2013scene} under dense settings (300 and 500 frames).
As shown, the proposed SAF3R consistently outperforms other efficient F3R methods for camera pose estimation across all three models under dense settings. Notably, the results are even close to those of the full-attention baselines, demonstrating the effectiveness of SAF3R in preserving critical 3D geometric relationships and cross-view correspondences during sparse attention computation.
For point cloud reconstruction, our method also achieves results close to the baseline models.

\subsubsection{Effectiveness of the DINO Top-$K$ kernel under sparse settings.}
\cref{tab:ablation dino topk vggt} and \cref{tab:ablation dino topk da3} present camera pose estimation results on the Co3D-v2 dataset~\cite{reizenstein2021common} under different key/value selection strategies for the VGGT and DA3 models, respectively. Similar to the results on 7Scenes reported in the main text, the DINO Top-$K$ selection strategy achieves comparable or higher accuracy than the uniform stride strategy for both models while using significantly fewer keys, indicating its effectiveness in retrieving highly activated keys from full attention. Notably, DINO Top-$K$ selection requires attending to less than 1\% of key tokens while recovering nearly the same accuracy as full attention, significantly reducing computational cost.

\begin{table}[h!]
  \centering
  \begin{minipage}[t]{0.49\textwidth}
    \centering
    \caption{
      Effectiveness of DINO Top-$K$ selection for VGGT on Co3D-v2 dataset. Sparse attention is applied only to heads $\{4, 10, 14\}$ in layer 14.
    }
    \label{tab:ablation dino topk vggt}
    \resizebox{\linewidth}{!}{
        \begin{tabular}{@{}lrccc@{}}
          \toprule
          {Kernel Type} 
          & {\#Keys} 
          & {AUC@30}$^\uparrow$ 
          & {AUC@3}$^\uparrow$ \\
          \midrule
          Full & 100\% & 90.34 & 58.26  \\
          \midrule
          Skip & 0\% & 19.55 & 0.03 \\
          \midrule
          Uniform Stride ($s$=64) & 1.6\% & 81.54 & 33.06 \\
          Uniform Stride ($s$=32) & 3.1\% & 86.85 & 43.77 \\
          \midrule
          DINO Top-$K$ ($K$=2) & 0.2\% & 87.44 & 45.34 \\
          DINO Top-$K$ ($K$=4) & 0.4\% & 88.05 & 48.05 \\
          \bottomrule
        \end{tabular}
    }
  \end{minipage}%
  \hfill
  \begin{minipage}[t]{0.49\textwidth}
    \centering
    \caption{
      Effectiveness of DINO Top-$K$ selection for DA3 on Co3D-v2 dataset. Sparse attention is applied only to heads $\{10, 17, 22\}$ in layer 7.
    }
    \label{tab:ablation dino topk da3}
    \resizebox{\linewidth}{!}{
        \begin{tabular}{@{}lrccc@{}}
          \toprule
          {Kernel Type} 
          & {\#Keys} 
          & {AUC@30}$^\uparrow$ 
          & {AUC@3}$^\uparrow$ \\
          \midrule
          Full & 100\% & 90.58 & 53.28 \\
          \midrule
          Skip & 0\% & 9.68 & 0.32 \\
          \midrule
          Uniform Stride ($s$=64) & 1.6\% & 89.31 & 51.12 \\
          Uniform Stride ($s$=32) & 3.1\% & 90.05 & 51.74 \\
          \midrule
          DINO Top-$K$ ($K$=2) & 0.2\% & 90.36 & 52.21 \\
          DINO Top-$K$ ($K$=4) & 0.4\% & 90.42 & 52.48 \\
          \bottomrule
        \end{tabular}
    }
  \end{minipage}
  
\end{table}

\begin{figure}[h!]
  \centering
  \includegraphics[width=\textwidth]{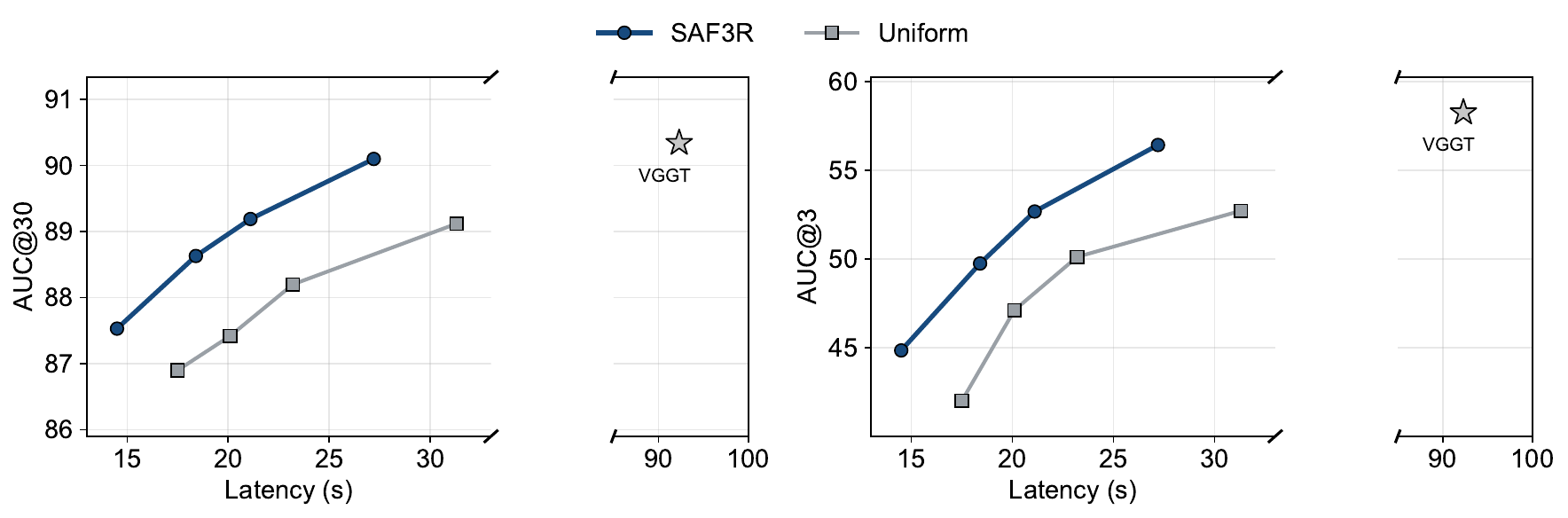}
  \caption{
  Camera pose estimation results on Co3D-v2 under different sparsity settings.
  }
  \label{fig:flexibility}
\end{figure}

\subsubsection{Flexibility of the proposed offline profiling method.}

Since the search procedure in the proposed offline profiling stage can start from any baseline sparsity level and progressively replace kernels to increase sparsity while maintaining accuracy, we can initialize the process with different sparsity baselines to obtain models with different lower-bounded sparsity levels. This flexibility allows the method to adapt to various deployment scenarios, such as devices with different computational and memory constraints.
\cref{fig:flexibility} presents camera pose estimation results on the Co3D-v2 dataset for the VGGT model using our method under different sparsity levels ranging from 75\% to 90\%, along with a uniform striding baseline at similar sparsity levels. As shown, by leveraging the heterogeneity of global attention patterns, our method outperforms the uniform striding baseline in camera pose estimation accuracy under similar latency. In addition, our method achieves performance comparable to the original VGGT model while being $3\times$ faster.

\subsubsection{Qualitative comparisons.}
\begin{figure}[h!]
  \centering
  \includegraphics[width=0.99\textwidth]{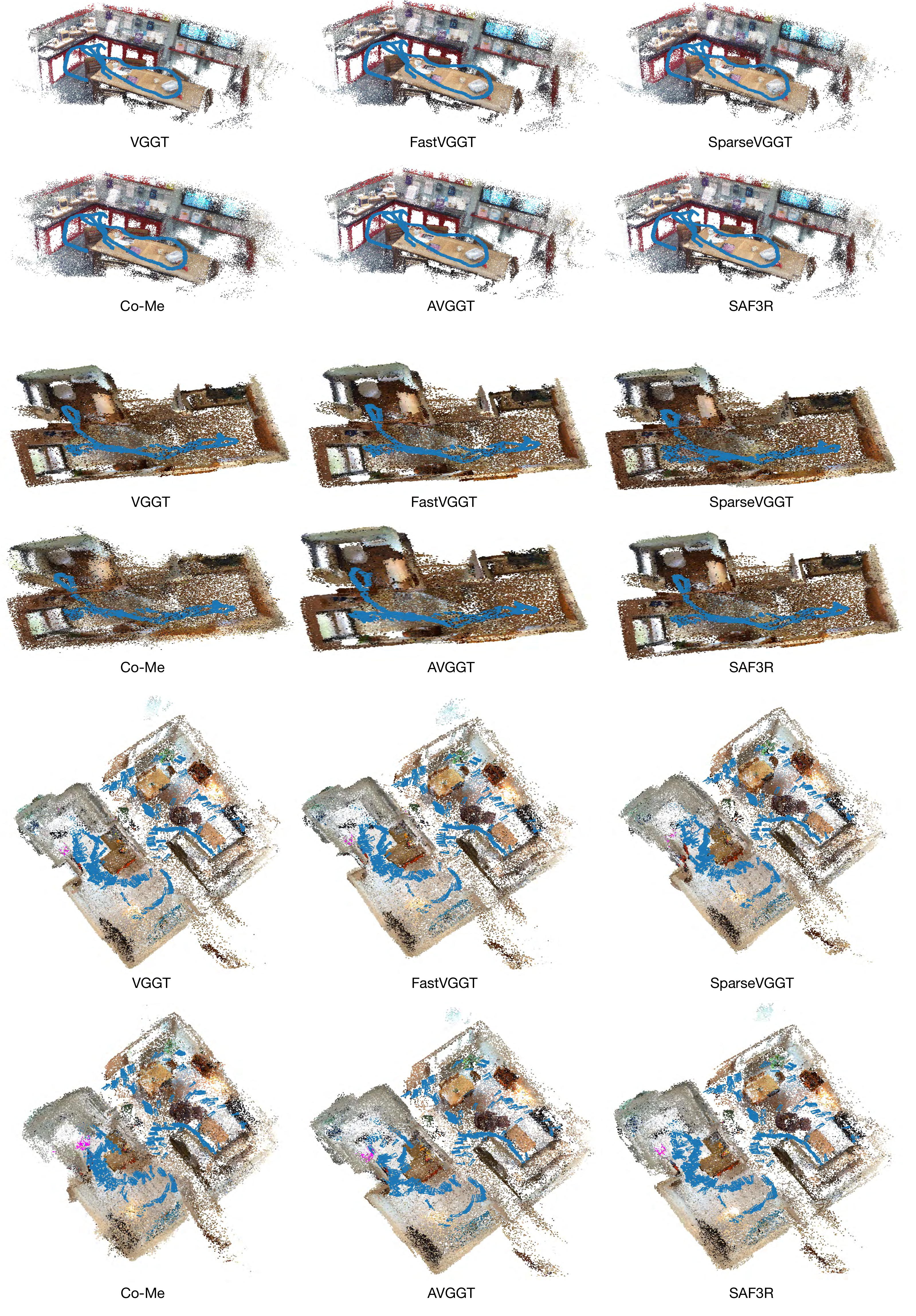}
  \caption{
  Qualitative comparisons of camera pose estimation and point cloud reconstruction for different efficient methods on VGGT. As shown, the proposed SAF3R method produces accurate camera poses and high-quality point clouds while achieving faster inference than other methods.
  }
  \label{appendix fig:qual cmp point clouds 1}
\end{figure}

\begin{figure}[h!]
  \centering
  \includegraphics[width=0.99\textwidth]{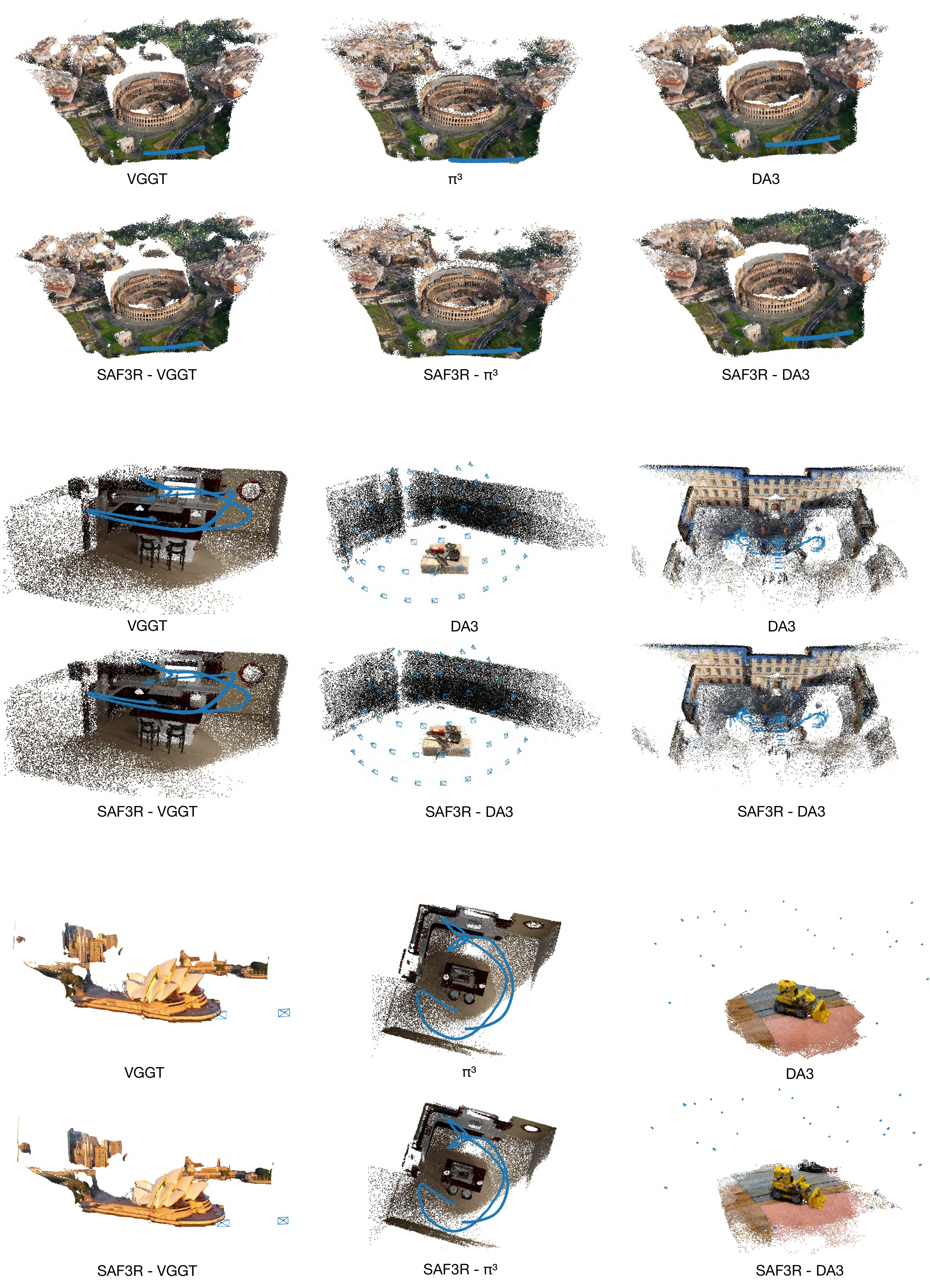}
  \caption{
  Qualitative comparisons of camera pose estimation and point cloud reconstruction between the original F3R transformers and our method. As shown, the proposed SAF3R demonstrates strong generality across different F3R models and effectively preserves the quality of the reconstructed point clouds.
  }
  \label{appendix fig:qual cmp point clouds 2}
\end{figure}

\cref{appendix fig:qual cmp point clouds 1} presents qualitative comparisons of estimated camera poses and reconstructed point clouds produced by different efficient methods on VGGT. As shown, the proposed SAF3R method produces accurate camera poses and high-quality point clouds while achieving faster inference than competing methods.
\cref{appendix fig:qual cmp point clouds 2} further compares the proposed method with the corresponding original F3R models. The examples are drawn from multiple datasets under different settings, ranging from sparse inputs with as few as 2 frames to dense sequences with up to 700 frames. As shown, the proposed SAF3R demonstrates strong generality across different F3R models and effectively preserves the quality of the reconstructed point clouds.

\end{document}